\documentclass[11pt]{article}
\pdfoutput=1

\usepackage[T1]{fontenc}
\usepackage[utf8]{inputenc}
\usepackage{hyperref}      
\usepackage{url}            
\usepackage{booktabs}       
\usepackage{amsfonts}       
\usepackage{nicefrac}       
\usepackage{microtype}      
\usepackage{graphicx}
\usepackage{amsmath,amssymb,amsthm}
\usepackage{float}
\usepackage{caption}
\usepackage{mathtools}
\usepackage{graphicx}
\usepackage{setspace}
\usepackage{lmodern}
\usepackage{xcolor}
\usepackage{enumitem}
\usepackage{placeins}
\usepackage{afterpage}

\usepackage{geometry}
\numberwithin{equation}{section}
\theoremstyle{plain}
\newtheorem{theorem}{Theorem}[section]
\newtheorem{lemma}[theorem]{Lemma}
\newtheorem{cor}[theorem]{Corollary}
\theoremstyle{definition}
\newtheorem{definition}{Definition}
\newtheorem{remark}{Remark}

\usepackage{times}

\usepackage{graphicx} 
\usepackage[numbers]{natbib}

\usepackage{algorithm}
\usepackage{algorithmic}
\usepackage{bbm}
\usepackage[caption=false]{subfig}

\def\argmin{\mathop{\rm arg\,min}}

\newcommand\blfootnote[1]{
  \begingroup
  \renewcommand\thefootnote{}\footnote{#1}
  \addtocounter{footnote}{-1}
  \endgroup
}

\title{Estimating Network Structure from Incomplete Event Data}

\author{Benjamin Mark$^{1}$, Garvesh Raskutti$^{1}$ and Rebecca Willett$^{2}$\\
$^1$University of Wisconsin-Madison \\ $^2$University of Chicago
}

\begin{document}
\maketitle

\begin{abstract}
Multivariate Bernoulli autoregressive (BAR) processes model time series of events in which the likelihood of current events is determined by the times and locations of past events.  These processes can be used to model nonlinear dynamical systems corresponding to criminal activity, responses of patients to different medical treatment plans, opinion dynamics across social networks, epidemic spread, and more.  Past work examines this problem under the assumption that the event data is complete, but in many cases only a fraction of events are observed. Incomplete observations pose a significant challenge in this setting because the unobserved events still govern the underlying dynamical system.  In this work,  we develop a novel approach to estimating the parameters of a BAR process in the presence of unobserved events via an unbiased estimator of the complete data log-likelihood function.  We propose a computationally efficient estimation algorithm which approximates this estimator via Taylor series truncation and establish theoretical results for both the statistical error and optimization error of our algorithm.  We further justify our approach by testing our method on both simulated data and a real data set consisting of crimes recorded by the city of Chicago. \blfootnote{This research is funded by the National Geospatial-Intelligence Agency's Academic Research Program (Award No. HM0476-17-1-2003, Project Title: Inference for Autoregressive Point Processes).
Approved for public release, 18-935.

This work was also supported by NSF CCF-1418976, NIH 1 U54 AI117924-01, NSF CCF-0353079, ARO W911NF-17-1-0357, and AFOSR FA9550-18-1-0166.
}\end{abstract}

\section{Introduction}

Discrete event data arises in a variety of forms including crimes, health events, neural firings, and social media posts.  Frequently each event can be associated with a node in a network, and practitioners aim to learn the relationships between the nodes in the network from the event data.  For example, one might observe a sequence of crimes associated with different gangs and seek to learn which crimes are most likely to spark retaliatory violence from rival gangs.

Such problems have attracted widespread interest in recent decades and a variety of point process models have been proposed to model such data.  The Hawkes process (\cite{hawkes1}) is a popular continuous time framework which has been applied in a number of different contexts (e.g., \cite{zhouZhaSongHawkes, yang_hawkes, clipped_hawkes, xu_hawkes}).  In addition, other works works have used a discrete time framework to model time series event data (e.g., \cite{lindermanneuro, fletcher_calcium,adverseDrugMLHC}).

A central assumption of many of these works is that {\em all} the events are observed.  However, in many cases we may observe only a subset of the events at random.  For example, while point process models have been widely used to model crime incidence (\cite{mohler2014hotspot, bertozzi2016field, linderman2014discovering}), frequently one only has access to {\em reported} crime data.  For many crimes the true number of incidents can be substantially higher.  The gap between the reported and true crime rates is referred to as ``the dark figure of crime'' by researchers in Sociology and Criminology who have studied this issue extensively (\cite{crime_statistics2, crime_statistics1}).  Unobserved events also pose a challenge in inference from Electronic Health Record (EHR) data which can be incomplete for a number of different reasons (\cite{ehr_missing1,ehr_missing2}). 

The unobserved events still play a role in the dynamical system governing the time series, making network estimation with incomplete data particularly challenging. In this paper, we contribute to the growing literature on modeling in the presence of unobserved events by proposing a novel method for estimating networks when we only observe a subset of the true events. 

\subsection{Problem Formulation}
\label{sec:formulation}
Many point process models of time series of discrete events are temporally discretized either because event times were discretized during the data collection process or for computational reasons. In such contexts, the temporal discretization is typically such that either one or zero events are observed in each discrete time block for each network node. With this in mind, we model the true but unobserved observations $X_1,\ldots X_T$ using a Bernoulli autoregressive process:
\begin{align}
    Y_t& =\nu+A^\ast X_{t-1}\notag \\
    X_t& \sim \mbox{Bernoulli} \left(\frac{1}{1+\exp(-Y_t)}\right). \label{eq:underlying_process}
\end{align}
Here $X_t \in \{0,1\}^M$ is a vector indicating whether events occurred in each of the $M$ nodes during time period $t$.  The vector $\nu \in \mathbb{R}^M$ is a constant bias term, and the matrix $A^\ast \in \mathbb{R}^{M \times M}$ is the weighted adjacency matrix associated with the network we wish to estimate.  
We assume that each row $a$ of $A^\ast$ lies in the $\ell_1$ ball of radius $r$, which we denote $\mathbb{B}_1(r)$.  We generally consider a case where $a$ is sparse and the magnitude of all its entries are bounded, so that $r$ is a universal constant which is independent of $M$. 

We observe $Z_1,\ldots Z_T$, a corrupted version of \eqref{eq:underlying_process} where only a fraction $p \in (0,1]$ of events are observed as follows:
\def\siid{\overset{iid}{\sim}}
\begin{align}
    W_{t,i}& \siid \mbox{Bernoulli}(p) \notag\\
    Z_t& =W_t \odot X_t.
    \label{eq:observed_process}
\end{align}
Here $\odot$ denotes the Hadamard product and $W_t \in \{0,1\}^M$ is a vector where each entry is independently drawn to be one with probability $p$ and zero with probability $1-p$.

Our analysis of \eqref{eq:underlying_process} and \eqref{eq:observed_process} can be naturally extended to several more complex variants.  Instead of assuming each $X_{t,i}$ is observed with probability $p$, we can assume events from each node $i$ are observed with a unique probability $p_i$.  We consider only a first order AR(1) process but our framework can be extended to incorporate more sophisticated types of memory as in \cite{mark_clipped}.  We omit discussion of these extensions in the interest of clarity.

\section{Related Work}

Corrupted or missing data in high-dimensional data sets is a problem which appears in a number of different domains and has attracted widespread interest over the past few decades (see \cite{missing_review} for an application-focused overview).  Our focus is on a particular type of corrupted data: partially observed sequences of discrete events.  In recent years researchers have started to focus on this problem (\cite{hawkes_missing_1, hawkes_missing_2, hawkes_missing_3}).  The prior works of which we are aware use a Hawkes process framework and assume knowledge of the time periods when the data is corrupted.  In the context of \eqref{eq:observed_process} this amounts to knowledge of $W_1,\ldots,W_T$.  Our method can operate in a setting where the researcher cannot differentiate between periods when no event actually occurred, and when events potentially occurred but were not recorded.  Moreover, because we use a discrete-time framework, we are able to derive sample complexity bounds for the estimation procedure proposed in Section \ref{sec:estimation}. Our theoretical results complement the empirical nature of much of the past work in this area.

This paper is also related a variety of works on regularized estimation in high-dimensional statistics, including \cite{BiRiTsy08, RasWaiYu10b, BasuMichailidis15} and \cite{jalali_missing}.  Many of these works have derived sample complexity guarantees using linear models, and some of these results have been extended to autoregressive generalized linear models (\cite{Neg10, hall2016inference}).  Another line of research (\cite{loh2011high, fast_global, loh_nonconvex, loh_linear, Neg10}) has formalized a notion of \textit{Restricted Strong Convexity} (RSC) which we leverage in Section \ref{sec:rates_opt}.  While many loss functions of interest in high-dimensional statistics are not strongly convex, these works have shown that they frequently satisfy an RSC condition which is sufficient to provide statistical and optimization error guarantees.  The main technical challenges in our setting lie in establishing results similar to these RSC conditions.  

\subsection{Missing Data in a High-Dimensional Linear Model}
\label{sec:linear}
\cite{loh2011high} straddles  the missing data literature and high-dimensional statistics literature.  The authors consider a missing data linear model
\begin{gather}Y_i=X_i^\top \beta^\ast + \epsilon_i \notag \\
Z_i=W_i \odot X_i \label{eq:l-w_model}
\end{gather} 
where $W_i \siid \mbox{Bernoulli}(p)$ and one observes pairs $(Y_i,Z_i)$ and aims to estimate $\beta^\ast$. 
The authors propose minimizing a loss function $L_{\text{missing},Z}$ of the observed data $Z$ which satisfies the property
$$\mathbb{E}[L_{\text{missing},Z,W}(\beta)|X]=L_{\text{Lasso},X}(\beta)$$ 
for any $\beta$. 
Here
$$L_{\text{Lasso},X} := \frac{1}{2}\sum_{i=1}^T (Y_i - X_i^\top \beta)^2 + \lambda \|\beta\|_1$$ denotes the classical Lasso loss function with the unobserved data $X$ and regularization parameter $\lambda>0$.  In other words, the missing data loss function is an unbiased estimator for the full data Lasso loss function we would ideally like to minimize.  This idea motivates our construction of a loss function for the observed process \eqref{eq:observed_process}.

Our problem can be viewed as an extension of \cite{loh2011high} to {\em autoregressive GLM models without knowledge of $W$}.\footnote{Note that \cite{loh2011high} does consider AR processes, but in a different context from our setting.  Specifically, we wish to estimate the AR process parameters, where as they  consider a special case of \eqref{eq:l-w_model} where $X_{t+1}=AX_t+\epsilon_t$ but where $A$ is known and one  aims to estimate $\beta^\ast$.}  In particular, we cannot distinguish events that were missed ($X_{t,j}=1$ and $W_{t,j}=0$) from correctly observed periods with no events ($X_{t,j}=0$).  \cite{loh2011high} are able prove sample complexity bounds as well as optimization bounds which are consistent with the high-dimensional statistical literature in that they scale with $\|\beta^\ast\|_0$ rather than the dimension of $\beta^\ast$.  We are able to prove analogous bounds for our estimator in Section \ref{sec:rates}.  

\subsection{Contributions}
This paper makes the following contributions.
\begin{itemize} 

\item We propose a novel method for estimating a network when only a subset of the true events are observed.  In contrast to previous work, our methods do not rely on knowledge of when the data is potentially missing.  Our procedure uses Taylor series approximations to an unbiased loss function, and  we show that these approximations have controlled bias and lead to accurate and efficient estimators.  

\item We prove bounds on both the statistical error and optimization error of our proposed estimation method.  The results hinge on showing that our loss function satisfies a restricted strong convexity (RSC) condition.  Past work on linear inverse problems with corrupted designs also establish RSC conditions, but these conditions do not carry over to the autoregressive GLM setting.  

\item We demonstrate the effectiveness of our methodology on both simulated data and real crime data.

\end{itemize}

\section{Proposed Estimation Procedure}
\label{sec:estimation}
Given the full data $X=[X_1,\ldots,X_T]$,  the negative log-likelihood function $L_X(A)$ is decomposable in the $M$ rows of $A$.  In other words, if $$A=\begin{bmatrix}a_1^\top &  a_2^\top & \cdots & a_M^\top  \end{bmatrix}^\top$$ then $L_X(A)=\sum_{m=1}^M L_X(a_m)$ where $a_m$ is the $m$th row of $A$ and $L_X$ denote the loss function restricted to a specific row.  Throughout the paper we slightly abuse notation and let $L_X(A)$ refer to the entire loss function when $A$ is a matrix, and let $L_X(a_m)$ refer to the loss function for a specific row when $a_m$ is a row vector.  The loss function for the $m$th row takes the form 
$$L_X(a_m):=\frac{1}{T} \sum_{t=1}^T f(a_m^\top X_t)-X_{t+1,m}(a_m^\top X_t)$$ where $f(x)=\log(1+\exp(x))$ is the partition function for the Bernoulli GLM.

We do not have access to $X$ and instead we aim to estimate $A$ using the corrupted data $Z=[Z_1,\ldots,Z_n]$.  As discussed in Section \ref{sec:linear}, our strategy will be to construct a loss function of $Z$ which is an unbiased estimator for $L_X$.  In other words, we want to find some function $L_{Z,p}$ such that for any $a \in \mathbb{B}_1(r)$,
\begin{equation}\mathbb{E}[L_{Z,p}(a_m)|X]=L_X (a_m).\label{eq:expectation_property}\end{equation}
In contrast to the Gaussian case discussed in Section \ref{sec:linear}, the Bernoulli partition function $f(x)=\log(1+\exp(x))$ is not a polynomial and constructing a function satisfying \eqref{eq:expectation_property} directly is challenging.  We adopt a strategy of computing unbiased approximations to truncated Taylor series expansions of $L_X$ and arriving at $L_{Z,p}$ as a limit of such approximations.  

To do this, we first rewrite $f$ using its Taylor series expansion around zero
\begin{align*}f(a_m^\top X_t)=\log(2)+\frac{a_m^\top X_t}{2}+\frac{(a_m^\top X_t)^2}{8}-\frac{(a_m^\top X_t)^4}{192}+o((a_m^\top X_t)^6).\end{align*}
The constant factor $\log(2)$ does not effect estimation in any way so we ignore it for the remainder of our discussion in the interest of simplicity.  We let $L_X^{(q)}$ denote the degree $q$ Taylor truncation to $L_X$.  The $X_t$ are binary vectors and we assume each row $a_m$ is sparse, so $a_m^\top X_t \leq \|a_m\|_1$ will not be too far from zero.  Thus it is reasonable to hope that for small $q$, $L_X^{(q)}(a_m)$ is a good approximation for $L_X(a_m)$ whenever $a_m \in \mathbb{B}_1(r)$.  We bound the approximation error in Lemma \ref{lemma:truncation_x} in the supplement. 

We now consider the problem of constructing a function $L_{Z,p}^{(q)}$ such that
\begin{equation}\mathbb{E}[L_{Z,p}^{(q)}(a_m)|X]=L_X^{(q)}(a_m) \text{ for all } a_m \in \mathbb{B}_1(r).\label{eq:desired_property_2}\end{equation}
Once we construct $L_{Z,p}^{(q)}$ we can estimate the $m$th row of $A^\ast$ by attempting to solve 
$$\hat a_m = \argmin_{a \in \mathbb{B}_1(1)} L_{Z,p}^{(q)}(a) + \lambda \|a\|_1.$$
Key question we need to address with this approach include (a) can we (approximately) solve this optimization problem efficiently? (b) will the solution to this optimization problem be robust to initialization? (c) will it be a strong estimate of the ground truth?

\subsection{Definition of \texorpdfstring{$L_{Z,p}^{(2)}$}{LZ}}
\label{sec:degree_two}
We first derive an unbiased estimator of the degree-two Taylor series expansion $L_X^{(2)}(a_m)$.
$$L_X^{(2)}(a_m)=\frac{1}{T} \sum_{t=1}^T \frac{a_m^\top X_t}{2}-X_{t+1,m}(a_m^\top X_t)+\frac{(a_m^\top X_t)^2}{8}.$$
Note that there are straightforward unbiased estimates of the first two terms:
\begin{align}\mathbb{E}\left[\frac{1}{p}\frac{a_m^\top Z_t}{2}|X\right]&=\frac{a_m^\top X_t}{2} \notag \\ \mathbb{E}\left[\frac{1}{p^2}Z_{t+1,m}(a_m^\top Z_t)|X\right]&=X_{t+1,m}(a_m^\top X_t).\label{eq:L_{Z,p}^21}\end{align}
For the third term, $\frac{(a_m^\top X_t)^2}{8}=\sum_{i,j} a_{m,i}a_{m,j}X_{t,i}X_{t,j}$, note that

\begin{equation}\mathbb{E}[Z_{t,i}Z_{t,j}|X]=\begin{cases} p^2 X_{t,i}X_{t,j} & \text{ if } i \not = j \\ p X_{t,i}X_{t,j} & \text{ if } i=j
\end{cases}.\label{eq:L_{Z,p}^22}\end{equation} 
Thus we must estimate the monomials with repeat terms ($i=j$) differently from the monomials with all distinct terms ($i \not =j$).
Using Equations \eqref{eq:L_{Z,p}^21} and \eqref{eq:L_{Z,p}^22} we can define the degree two unbiased estimator:
\begin{align}L_{Z,p}^{(2)}(a_m):=\frac{1}{T} \sum_t \Bigg[ \frac{a_m^\top Z_t}{2p}-\frac{Z_{t+1,m}(a_m^\top Z_t)}{p^2} +\sum_{i \not =j} \frac{a_{m,i}a_{m,j}Z_{t,i}Z_{t,j}}{8p^2} +\sum_i \frac{a_{m,i}^2Z_{t,i}}{8p}\Bigg]. \label{eq:L_{Z,p}^2}
\end{align}

\subsection{Higher-Order Expansions}

The construction of $L_{Z,p}^{(2)}$ in the previous section suggests a general strategy for constructing  $L_{Z,p}^{(q)}$ satisfying \eqref{eq:desired_property_2}.  Take any monomial $$X_{t,m_1}\cdot \ldots \cdot X_{t,m_d}$$
depending on the counts in nodes $m_1,\ldots m_d$ during time period $t$.   Wherever this monomial appears in $L_X^{(q)}(a_m)$, our unbiased loss function will have a term $$\frac{1}{p^k}Z_{t,m_1}\cdot \ldots \cdot Z_{t,m_d}$$ scaled by $\frac{1}{p^k}$ where $k$ denotes the number of unique terms in the monomial.  For example, in Equation \eqref{eq:L_{Z,p}^21} each degree two monomial was unique so we scaled everything by $\frac{1}{p^2}$.  However, in \eqref{eq:L_{Z,p}^2} some of the degree two monomials had repeated terms and so they were scaled by $\frac{1}{p}$. In order to formalize these ideas and generalize our estimator to $q>2$, we first need to introduce additional notation.

\subsubsection{Notation}

  Let $\mathcal{U}_d$ denote the set of all monomials of degree $d$.  We represent an element $U \in \mathcal{U}_d$ as a list containing $d$ elements.  An element in the list corresponds to the index of a term in the monomial.  For an example, the monomial $x_1^2x_3$ can be represented as the list $(1,1,3)$.  

For a polynomial function $h$ we let $c_{U,h}$ denote the coefficient of the monomial $U$ in $h$.  Finally we define the order of a list to denote the number of unique elements in the list, so $|(1,2)|=2$ whereas $|(1,1)|=1$.    

\paragraph{Example}\textit{Consider the function $h(x_1,x_2)=x_1^2+4x_1x_2$.  We can decompose $h$ as 
$$h(x_1,x_2)=\sum_{U \in \mathcal{U}_2} c_{U,h} \prod_{u \in U} x_u$$ where $\mathcal{U}_2=\{(1,1), (1,2),(2,2)\}$ with corresponding coefficients $c_{(1,1),h}=1$, $c_{(1,2),h}=4$ and $c_{(2,2),h}=0$.}  

Using this notation we can write \begin{align*}
L_X^{(q)}(a_m)=\frac{1}{T} \sum_{t=1}^T \Bigg[\sum_{i=1}^q \sum_{U \in \mathcal{U}_i} \left( c_{U,f} \prod_{u \in U} X_{t,u}\prod_{u \in U}a_{m,u} \right)
-X_{t+1,m}(a_m^\top X_t)\Bigg].
\end{align*}

\subsubsection{Definition of \texorpdfstring{$L_{Z,p}^{(q)}$}{L}}
The degree $q$ likelihood is constructed as follows \begin{align*}
L_{Z,p}^{(q)}(a_m):=\frac{1}{T} \sum_{t=1}^T \Bigg[\sum_{i=1}^q \sum_{U \in \mathcal{U}_i} \left(\frac{c_{U,f}}{p^{|U|}} \prod_{u \in U} Z_{t,u}\prod_{u \in U}a_{m,u} \right)
-\frac{Z_{t+1,m}(a_m^\top Z_t)}{p^2}\Bigg].\end{align*}
Recall that $|U|$ denotes the number of unique terms in the monomial $U$.  In other words, in we adjust $L_X^{(q)}$ by scaling each monomial according the the number of unique terms rather than the number of overall terms.  This definition clearly satisfies \eqref{eq:desired_property_2}.  We show in the supplement that if $r=1$ and $p>\frac{1}{\pi}$ then $\lim_{q \to \infty} L_{Z,p}^{(q)}(a_m)$ converges uniformly on $\mathbb{B}_1(r)$ to a function we denote as $L_{Z,p}(a_m)$.  Extending this loss function on individual rows to an entire matrix, we can define $L_{Z,p}(A)=\sum_{i=1}^M L_{Z,p}(a_m)$.  
An additional technical discussion in the supplement guarantees that $L_{Z,p}$ actually satisfies the desired property in Equation \eqref{eq:expectation_property}.

 \subsection{Proposed Optimization}
 \label{sec:optimization}
 In practice we can only compute $L_{Z,p}^{(q)}$ for finite $q$.  To estimate $A^\ast$ we  consider the following constrained optimization:
\begin{equation}\widehat{A} \in \argmin_{A \in \mathbb{B}_{1,\infty}(r)} L_{Z,\hat{p}}^{(q)}(A)+\lambda \|A\|_1 \label{eq:estimator}\end{equation}
where $\hat{p}$ is an estimate of the missingness parameter $p$ and 
$$\mathbb{B}_{1,\infty}(r) = \{ A \in \mathbb{R}^{M \times M} : ||a_m||_1 \leq 1 \text{ for all } m\}.$$

In general, $L_{Z,p}^{(q)}$ is not a convex function.  However, we  show in Section \ref{sec:rates_opt} that under certain assumptions all stationary points of \eqref{eq:estimator} must lie near $A^\ast$.  Thus we can approximately solve \eqref{eq:estimator} via a simple projected gradient descent algorithm.

In order to apply our algorithm it is necessary to have an estimate $\hat{p}$ of the frequency of missed data.  In many cases one may have prior knowledge available.  For example, social scientists have attempted to quantify the frequency of unreported crimes (\cite{crime_statistics1, crime_statistics3}).  Moreover, a simulation study in Section \ref{sec:simulations} suggests our strategy is robust to misspecification of $p$.

\section{Learning Rates}
\label{sec:rates}
In this section we answer several questions which naturally arise with our proposed estimation procedure in Equation \eqref{eq:estimator}.  In particular, we'd like to know:
\begin{itemize}
    \item If we can find a solution to Equation \eqref{eq:estimator}, will it be a good approximation to $A^\ast$?
    \item The loss function $L_{Z,p}^{(q)}$ is not convex.  Is it possible to actually find a minimizer to \eqref{eq:estimator}, or an approximation to it?
    \end{itemize}
In Theorem \ref{thm:one} we answer the first question affirmatively.  In Theorem \ref{thm:two} we show that all stationary points of \eqref{eq:estimator} are near one another, so that simple gradient-based methods will give us a good solution. 

Throughout this section we assume $p>\frac{1}{\pi}$ and $A^\ast \in \mathbb{B}_{1,\infty}(1)$.  All results in the section apply for the loss functions $L_{Z,p}^{(q)}$ for $q \in \mathbb{N} \cup \{\infty\}$.  In the $q=\infty$ case we recover the idealized loss function $L_{Z,p}$.  

We use $a \lesssim b$ to mean $a \leq Cb$ and $a \asymp b$ to mean $a=Cb$ where $C$ is a universal constant.  Define $s:=\|A^\ast\|_0$ and $\rho:= \max_m \|a_m^\ast\|_0$. 

\subsection{Statistical Error}
\label{sec:rates_statistical}

The following theorem controls the statistical error of our proposed estimator.

\begin{theorem}[Accuracy of $L_{Z,p}^{(q)}$]

Suppose $$\widehat{A} \in \argmin_{A \in \mathbb{B}_{1,\infty}(1)} L_{Z,p}^{(q)}(A)+\lambda \|A\|_1$$ where $\lambda \asymp \frac{\log(MT)}{\sqrt{T} (p\pi-1)} + \frac{1}{(p\pi)^q}$.  Then $$\|\widehat{A}-A^\ast\|_F^2 \lesssim \frac{s\log^2(MT)}{T(\pi p-1)^2}+\frac{s}{(p\pi)^{2q}}$$ for $T \gtrsim \rho^2 \log(MT)$ with probability at least $1-\frac{1}{T}$.  
\label{thm:one}
\end{theorem}

The two terms in the upper bound of Theorem \ref{thm:two} have a natural interpretation.  The first represents the error for the idealized estimator $L_{Z,p}$, while the second represents the error due to the Taylor series truncation.  Our error scales as $(\pi p-1)^{-2}$ which is reasonable in the context of our algorithm because $L_{Z,p}(A):=\lim_{q \to \infty} L_{Z,p}^{(q)}(A)$ is only well-defined when $p>\frac{1}{\pi}$ (see Remark \ref{remark:technical} in the supplement). An interesting open question which arises from Theorem \ref{thm:one} is whether the process described in  \eqref{eq:underlying_process} and \eqref{eq:observed_process}
is unidentifiable for $p \leq \frac{1}{\pi}$ or  something specific to our methodology fails for $p$ below this threshold.  

The proof of Theorem \ref{thm:one} uses ideas from the analysis of high-dimensional GLMs (\cite{hall2016inference, mark_clipped}) as well as ideas from the analysis of missing data in the linear model \cite{loh2011high} and Gaussian linear autoregressive processes \citep{jalali_missing}.  The key technical challenge in the proof lies in controlling the gradient of the error term $R^{(q)}(A):= L_X(A)-L_{Z,p}^{(q)}(A)$.  This is done in Lemmas \ref{lemma:prelim}-\ref{lemma:error2} in the supplement.

\subsection{Optimization Error}
\label{sec:rates_opt}
We next focus on the optimization aspects of Equation \eqref{eq:estimator}.  Our loss function $L_{Z,p}^{(q)}$ is non-convex, so at first glance it may appear to be a poor proxy loss function to optimize.  However, a body of research (see \cite{fast_global, loh_nonconvex, loh2011high, loh_linear}) has studied loss functions satisfying a properties known as restricted strong convexity (RSC) and restricted smoothness (RSM).  These works have shown that under certain conditions, the optimization of non-convex loss functions may be tractable.  The formal definitions of the RSC and RSM conditions we use are as follows.

\begin{definition}[Restricted Strong Convexity]Let $T_L(v,w)$ denote the first order Taylor approximation to a loss function $L$ centered at $w$.  A loss function $L$ satisfies the RSC condition with parameters $\alpha,\tau$ if 
$$T_L(v,w) \geq \frac{\alpha}{2}\|v-w\|_2^2-\tau \|v-w\|_1^2$$ for all $v,w \in \mathbb{B}_1(1)$. 
\end{definition}
\begin{definition}[Restricted Smoothness]
A loss function $L$ satisfies the RSM condition with parameters $\alpha,\tau$ if 
$$T_L(v,w) \leq \frac{\alpha}{2}\|v-w\|_2^2+\tau \|v-w\|_1^2$$ for all $v,w \in \mathbb{B}_1(1)$. 
\end{definition}

We are able to show these conditions are satisfied for $\alpha \asymp 1$ and $\tau \asymp \sqrt{\frac{\log(MT)}{T}}+(p\pi)^{-q}$.  This in turn gives the following result.  As in Theorem \ref{thm:one} we assume $p>\frac{1}{\pi}$.

\begin{theorem} 
 
Suppose $A^\ast \in \mathbb{B}_{1,\infty}(1)$ and $\|a_m^\ast\|_0>0$ for at least $\frac{M}{C}$ rows of $A^\ast$ where $C$ is a universal constant.  Let $\tilde{A} \in \mathbb{B}_{1,\infty}(1)$ be any stationary point of $L_{Z,p}^{(q)}(A)+\lambda \|A\|_1$ where $\lambda \asymp \frac{\log(MT)}{\sqrt{T} (p\pi-1)}+\frac{1}{(p\pi)^q}$.  Then 
$$\|\tilde{A}-A^\ast\|_F^2 \lesssim  \frac{s}{p \pi-1}\sqrt{\frac{\log(MT)}{T}}+\frac{s}{(p\pi)^q} $$
with probability at least $1-\frac{\log(T)}{T^2}$ for $T \gtrsim \rho^2 \log(MT)$ and $q \gtrsim \frac{\log(\rho)}{\log(\pi p)}$.

\label{thm:two}
\end{theorem}

As in Theorem \ref{thm:one} the first term in our bound can be interpreted as the error for the idealized estimator $L_{Z,p}$ while the second term can be thought of as the error due to the Taylor series truncation.  The assumption that $\|a_m^\ast\|_0>0$ for at least $\frac{M}{C}$ rows of $A^\ast$ says that at least a constant fraction of nodes are influenced by other nodes in the network.  This assumption allows us to state Theorem \ref{thm:two} in terms of $s$ - the support of $A^\ast$.  In extreme cases where almost all nodes in the network fire independently of the other nodes it is possible for the optimization error to have a slower scaling than $s$.

The RSC and RSM conditions are closely related to ideas used in our statistical error bounds in Theorem \ref{thm:one}.  Lemma \ref{lemma:rsc/rsm} shows that the conditions are satisfied for $\tau$ on the order of $\frac{1}{\sqrt{T}}$ which leads to an overall optimization error bound of the same order.  This is a slower convergence rate than in the linear case; whether stronger rates can be obtained in the autoregressive GLM setting is an open question.

The proof of Theorem \ref{thm:two} proceeds as follows.  We first establish that the RSC/RSM conditions hold for reasonable constants in Lemma \ref{lemma:rsc/rsm}.  This proofs relies on the technical machinery built up in Lemmas \ref{lemma:prelim}-\ref{lemma:error2}.  We can then combine our RSC/RSM results with Theorem 2 in \cite{fast_global} to conclude that all stationary points of $L_{Z,p}^{(q)}(A)+\lambda \|A\|_1$ lie in a small neighborhood of $A^\ast$ with high probability.

\section{Simulations}
\label{sec:simulations}

\afterpage{\begin{figure}[t!]
\centering    
\subfloat[$p=\hat{p}=.6$]{\label{fig:figure2_60}\includegraphics[width=65mm]{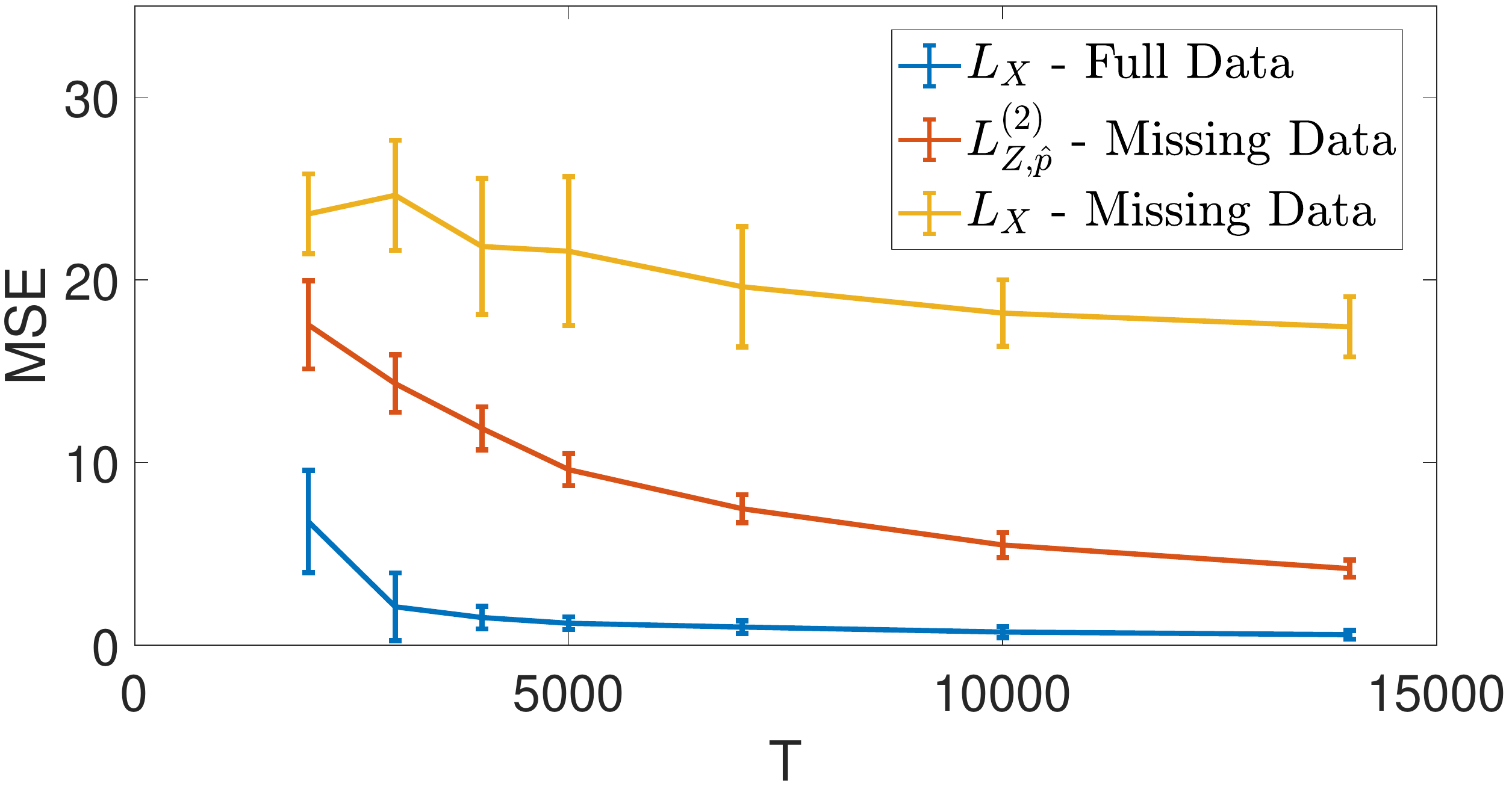}}~
	\subfloat[$p=\hat{p}=.75$]{\label{fig:figure2_75}\includegraphics[width=65mm]{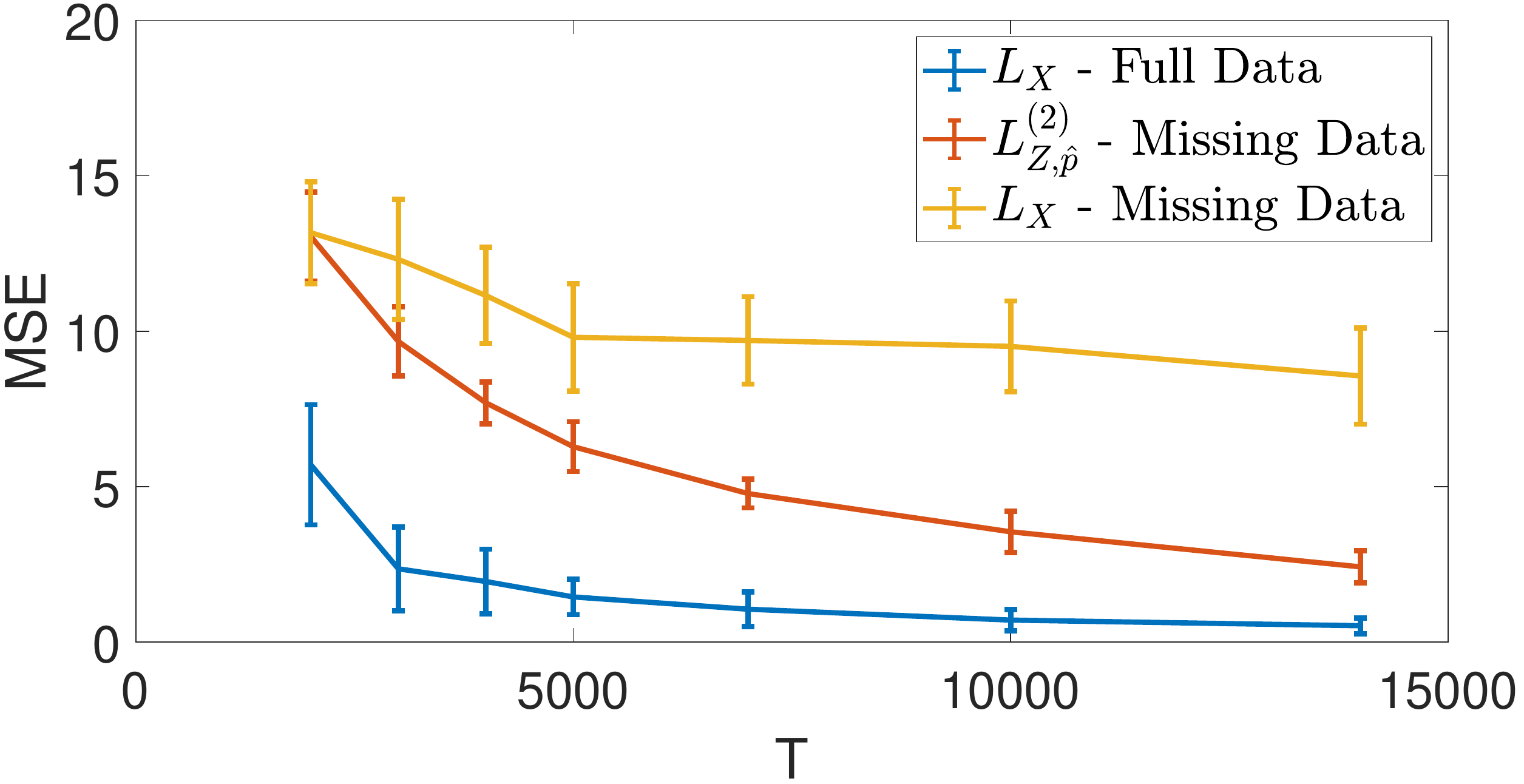}}
\caption{\small MSE vs $T$ for $p=.6$ (top) and $p=.75$ (bottom).  The blue line uses regularized MLE on full data -- i.e. data unavailable in our setup -- and represents a kind of oracle estimator.  The red line uses incomplete data with $L_{Z,p}^{(2)}$ (our proposed method).  The yellow lines corresponds to minimizing the full data likelihood over the incomplete data -- that is, this estimator naively ignores the issue of missing data.  Median of 50 trials is shown and error bars denote sample standard deviations.}

\label{figure:figure2} 
\end{figure}}

In this section we evaluate the proposed method on synthetic data.  We generate $A^\ast \in \mathbb{R}^{50 \times 50}$ with $s=50$ nonzero entries with locations chosen at random.  Each nonzero entry is chosen uniformly in $[-1,1]$ and $\nu=0$.  We then generate a ``true" data set $X$ and an ``observed'' data set $Z$ according \eqref{eq:underlying_process} and \eqref{eq:observed_process} with $\lambda=\frac{.75}{\sqrt{T}}$.  We perform projected gradient descent with a random initialization and show a median of 50 trials.

Figure \ref{figure:figure2} shows mean squared error (MSE) vs $T$ for $p=.6$ (top) and $p=.75$ (bottom).  Our method is shown in red.  It uses the loss function $L_{Z,p}^{(2)}$ on the partially observed data $Z$.  Our method is compared to the loss function $L_X$ using both the full data $X$ (i.e., an oracle estimator with access to the missing data) and the partially observed data $Z$ (i.e. a naive estimator that ignores the possibility of missing data).  As expected, with access to the complete data one can get a more accurate estimate of $A^\ast$ than either method using the partially observed data.  However, our method outperforms the full data likelihood when given the partially observed data.  In particular, note that the accuracy for the full data likelihood stalls after some time due to the inherent bias in using the corrupted data on the true data likelihood.  In contrast our unbiased method continues to converge to the solution, as suggested by the results in Section \ref{sec:rates}. Finally, observe that for large $T$ there is little variation between trials when using $L_Z^2$ even though each trial was initialized randomly.  This agrees Theorem \ref{thm:two} which states that all stationary points of $L_{Z,p}^{(q)}$ lie near one another.

In practical applications one may have strong reason to believe some events are unobserved, but pinning down a precise estimate of the missingness parameter $p$ might be unrealistic.  Therefore it is important to see how our algorithm performs as a function of the misspecification $\hat{p}-p$.  We examine this in Figure \ref{fig:robustness}.  We generate data as in the previous section but with $p=.7$.  We then apply our algorithm with the loss function $L_{Z,\hat{p}}^{(2)}$ and varying values of $\hat{p}$.  

\afterpage{\begin{figure}[t!]
\centering    
\includegraphics[width=85mm]{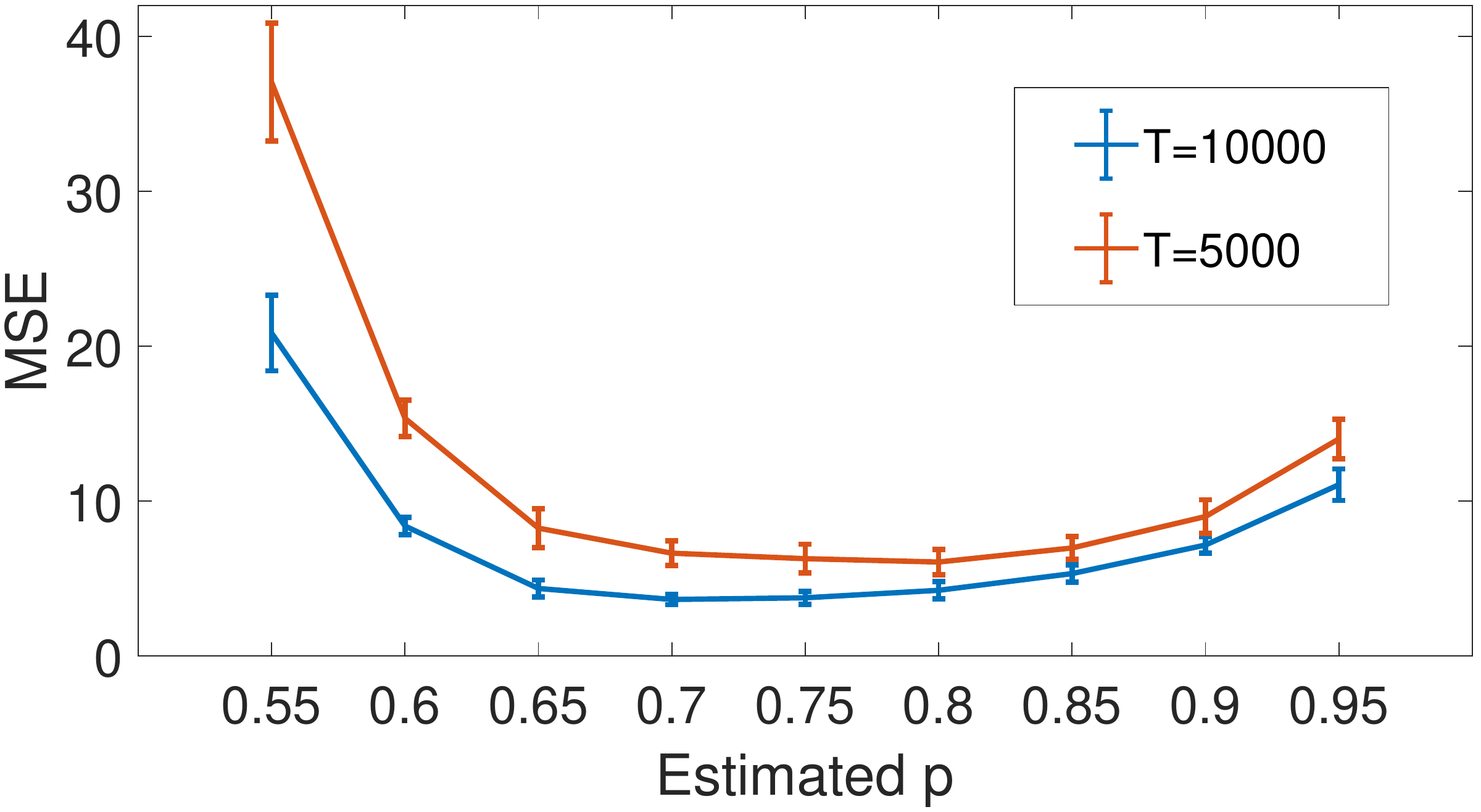}
\caption{\small Robustness to missestimation of $p$ using a true value of $p=.7$.  Median of 50 trials is shown and errorbars denote sample standard deviations.}
\label{fig:robustness}
\end{figure}}

Figure \ref{fig:robustness} shows that our method is highly robust to small misspecification of the missingness parameter $p$.  Interestingly, underestimating $p$ by more than 10\% leads to poor results but our method is still robust to overestimation of over 10\%. This suggests there is value in applying our techniques with a conservative estimate of the amount of missed data, even when one has only a rough estimate of the frequency of missed events.

As final experiment we measure how MSE varies as a function of the Taylor series truncation level $q$.  Calculating $L_{Z,p}^{(4)}$ takes a significant amount of time for high-dimensional problems, so we randomly generate $A^\ast \in \mathbb{R}^{20 \times 20}$ with $s=20$ nonzero entries compared to $50$ in the previous simulations and run $30$ trials.  We set $p=\hat p=.7$.

\afterpage{\begin{figure}
\centering    
\includegraphics[width=85mm]{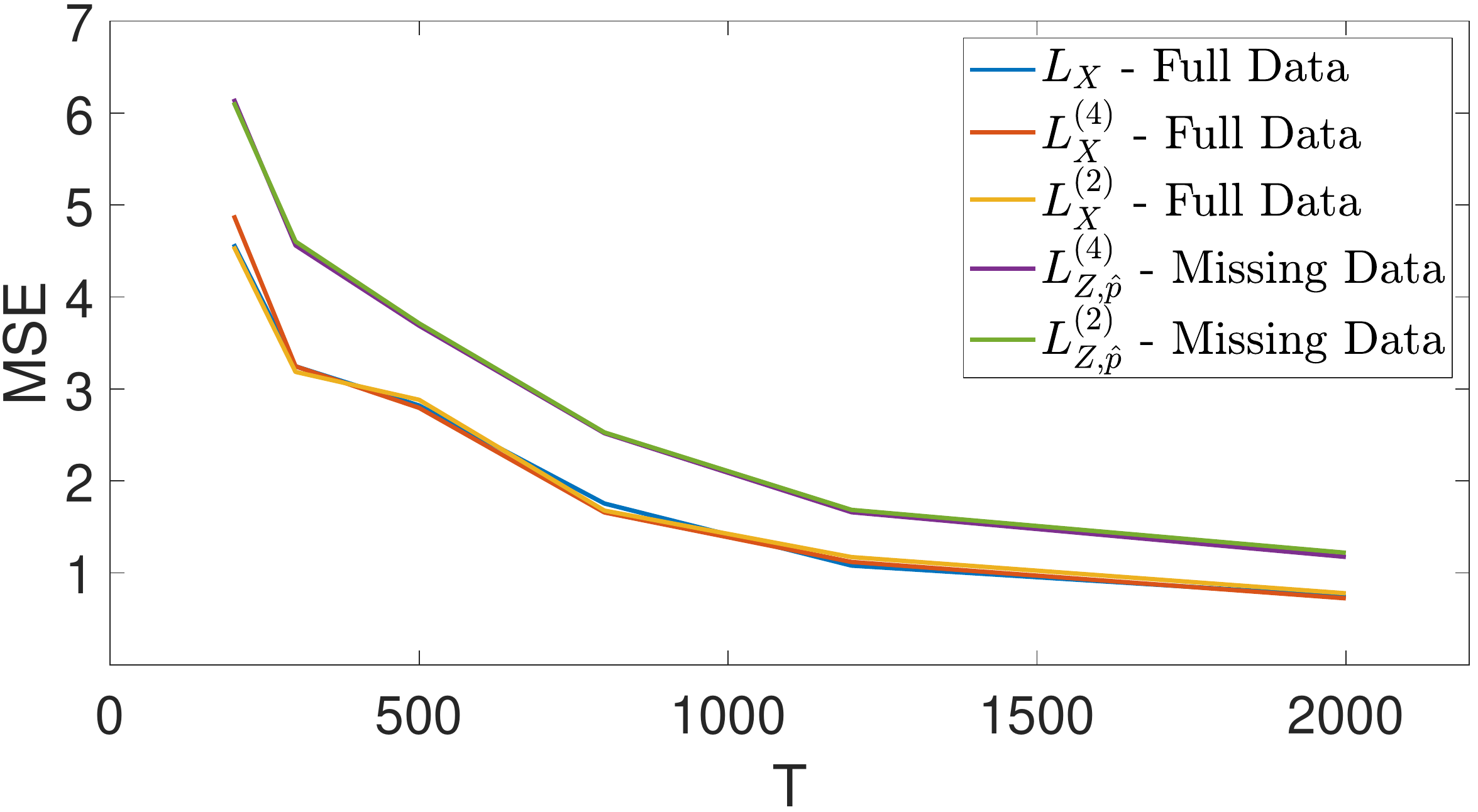}
\caption{\small MSE vs $T$ using different loss functions in \eqref{eq:estimator}.  $L_X, L_X^4$ and $L_X^2$ use the full data $X$ while $L_{Z,p}^{(4)}$ and ${(2)}$ use the missing data $Z$.  Plots suggest that Taylor series truncations produce nearly identical results to complete loss functions.}
\label{figure:figure1}
\end{figure}}

In Figure \ref{figure:figure1} we show MSE as a function of $T$ for the full data loss function at three different truncation levels: $L_X^2, L_X^4$ and $L_X$.  Recall that $L_X$ and $L_{Z,p}$ has no odd degree terms other than $1$, so $L_X^{(3)}=L_X^{(2)}$ and $L_{Z,p}^{(3)}=L_{Z,p}^{(2)}$.  We see that the second and fourth degree truncations perform essentially the same as the full data likelihood.  We also plot MSE as a function of $T$ for the truncated missing data loss functions $L_{Z,p}^{(2)}$ and $L_{Z,p}^{(4)}$.  As expected, using the full data gives stronger results than the partially observed data.  We again see that the second and fourth order truncations perform nearly the same.  The sample standard deviations are also similar - e.g. when $T=2000$ the standard deviations of $L_X, L_X^2$ and $L_X^4$ are $.184, .178$ and $.186$ respectively while the standard deviations for $L_Z^2$ and $L_Z^4$ are $.322$ and $.311$.  The similarity between the second and fourth order truncation levels suggests that choosing one of these truncation levels will give us a strong approximation to $L_{Z,p}$.  Since $L_{Z,p}^{(4)}$ takes significantly longer to compute, we use the second order approximation in the first two experiments.

\section{Chicago Crime Data}

This section studies a data set consisting of crimes committed in Chicago between January 2001 and July 2018 (\cite{citychicago}).  Point process models have been applied to this data set in the past (\cite{linderman2014discovering}).  In a missing data setting, in order to validate our model it is important to have a ``ground truth'' data set.  For this reason we limit our study to homicides within the data set.  For other crimes recorded data is known to be incomplete (\cite{crime_statistics1,crime_statistics2}), but we assume that nearly every murder is observed.  This allows us to create an incomplete data set by removing murders randomly while still maintaining a ground truth data set for validation. 

The city is divided into 77 community areas and the community area where each murder occurred is recorded.  The majority of these community areas experience few murders so we focus on the nine areas with the most murders since 2001.  These areas form two clusters: one on the west side of the city and another on the south side.  We discretize the events using one week bins, so $X_{t,i}=1$ if a murder occurred in community area $i$ during week $t$ and $X_{t,i}=0$ otherwise.  This gives a data matrix $X \in \{0,1\}^{9 \times 918}$ which we divide into a train set $X_{\text{train}} \in \{0,1\}^{9 \times 600}$ containing the first 600 weeks in the period, and a test set $X_{\text{test}} \in \{0,1\} ^{9 \times 318}$ containing the final 318 weeks.  We then create an incomplete data set $Z_{\text{train}}=W \odot X_{\text{train}}$ where $W \in \{0,1\}^{9 \times 318}$ contains independent realizations of a Bernoulli random variable with mean $p=.75$.  

We learn parameters $\nu_X \in \mathbb{R}^9$ and $\hat A_X \in \mathbb{R}^{9 \times 9}$ using the training set $X_{\text{train}}$ and the full data likelihood $L_X$. We also learn parameters $\nu_{Z,\hat{p}} \in \mathbb{R}^9$ and $\hat A_{Z,\hat{p}} \in \mathbb{R}^{9 \times 9}$ using the incomplete train data $Z_{\text{train}}$ and the missing data likelihood $L_{Z,\hat{p}}^2$ for various values of $\hat{p}$. 

\begin{figure}
\centering    
\includegraphics[width=85mm]{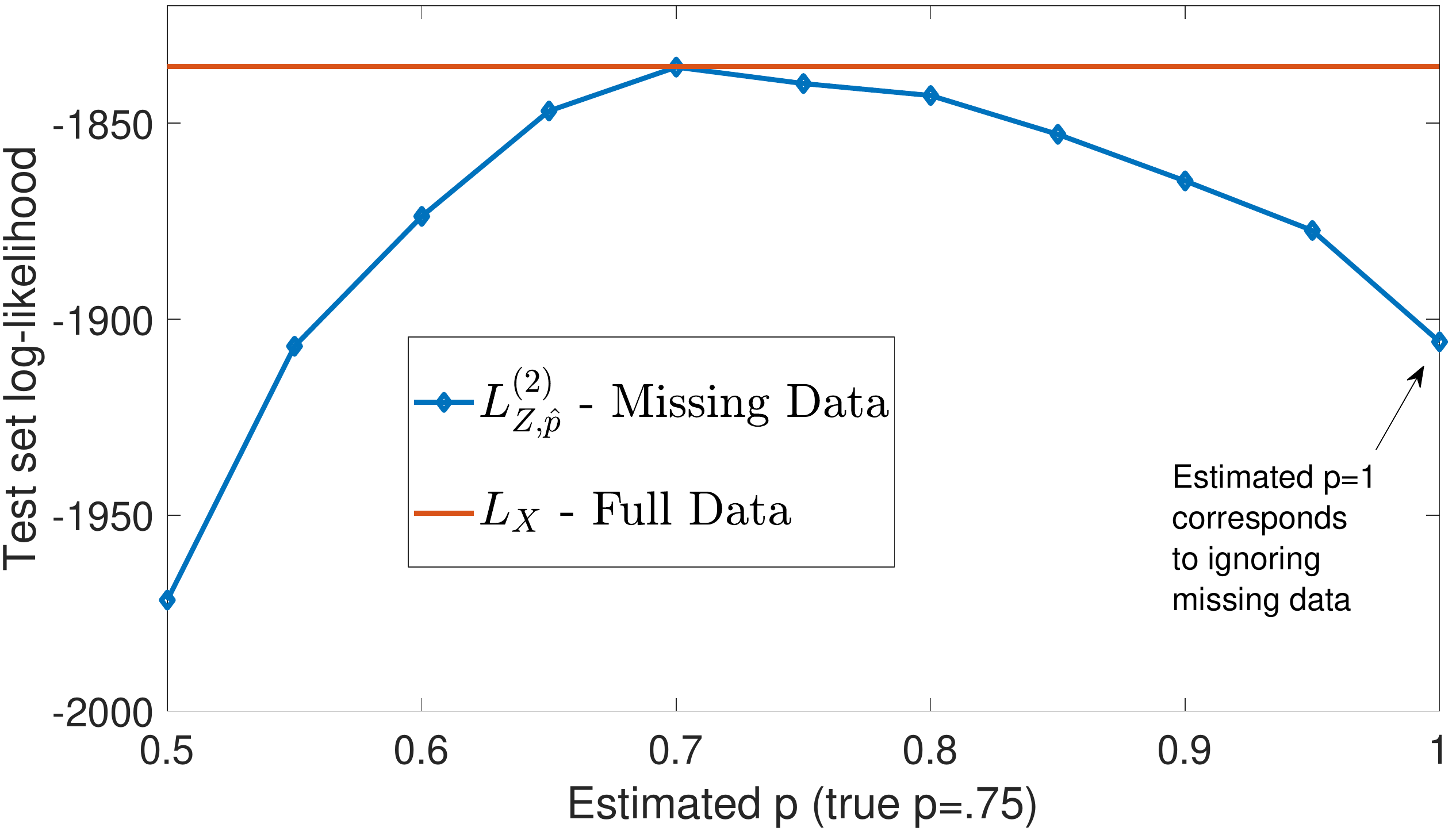}
\caption{\small Test performance on Chicago crime data. Log-likelihood of events on hold out set using full data with $L_X$ (yellow) and partial data with $L_{Z,\hat{p}}^2$ for various values of $\hat{p}$, where $p=0.75$ (blue). For $\hat p$ near $p$, the proposed estimator performs nearly as well as an oracle estimator with full access to the missing data, and significantly better than a naive method that ignores the possibility of missing data.
}
\label{figure:figure4}
\end{figure}

We compare the log-likelihood of these parameters on the test set $X_{\text{test}}$.  The results are shown in Figure \ref{figure:figure4}. The missing data estimates perform nearly as well as the full data estimate when $\hat{p}$ is close to the true value of $.75$.  Note that $L_{Z,1}^2=L_X^2$ closely approximates the full data likelihood $L_X$ and the hold out likelihood is substantially worse for $L_{Z,1}^2$ compared to $L_{Z,\hat{p}}^2$ for $\hat{p}$ close to $.75$.  In other words, ignoring the missing data entirely gives a weaker estimate than applying the techniques this paper introduces, even when $\hat{p}$ is not a perfect estimate of $p$.  Finally, observe that $L_{Z,\hat{p}}$ is more robust to misspecification when $\hat{p}>p$ compared to when $\hat{p}<p$.  This is a trend which also appears in Figure \ref{fig:robustness} and suggests there is value in using conservative estimates of the amount of missed data in practical applications.

Given estimates of $A$ and $p$ we can use density propagation to predict the likelihood of homicides during week $n$ based on observed homicides up to week $n-1$.  We do this for $\hat A_{Z,.75}$ learned from the incomplete data $Z_{\text{train}}$ with $\hat p =.75$ as well as $\hat A_{Z,1}$ learned from $Z_{\text{train}}$ but with $\hat p =1$, which corresponds to assuming there is no missing data.  We use particle filtering to construct estimates $$p(X_n=1 | \hat A_{Z,.75}, Z_1,\ldots Z_{n-1})$$
and $$p(X_n=1 | \hat A_{Z,1}, Z_1,\ldots Z_{n-1}).$$  These probabilities correspond to the likelihood of homicides during the $n$th week based on the observations over the first $n-1$ weeks.  We construct such estimates for each week in the $318$ week test set.  As expected $\hat A_{Z,.75}$ assigns higher likelihoods of homicides, with $960$ expected homicides in total compared to $748$ for $\hat A_{Z,1}$.  As a naive method of correcting for missing data, we divide $p(X_n=1 | \hat A_{Z,1}, Z_1,\ldots Z_{n-1})$ by a constant scaling factor of $0.75$ and report these likelihoods below; by doing this, we ensure that both predictions yield similar {\em average} numbers of homicides, so differences in performance between the proposed and naive estimator are not due to a simple difference in averages, but rather because the proposed method is capturing the underlying dynamics of the system.  

\begin{figure}[!ht]
\centering    
\includegraphics[width=100mm]{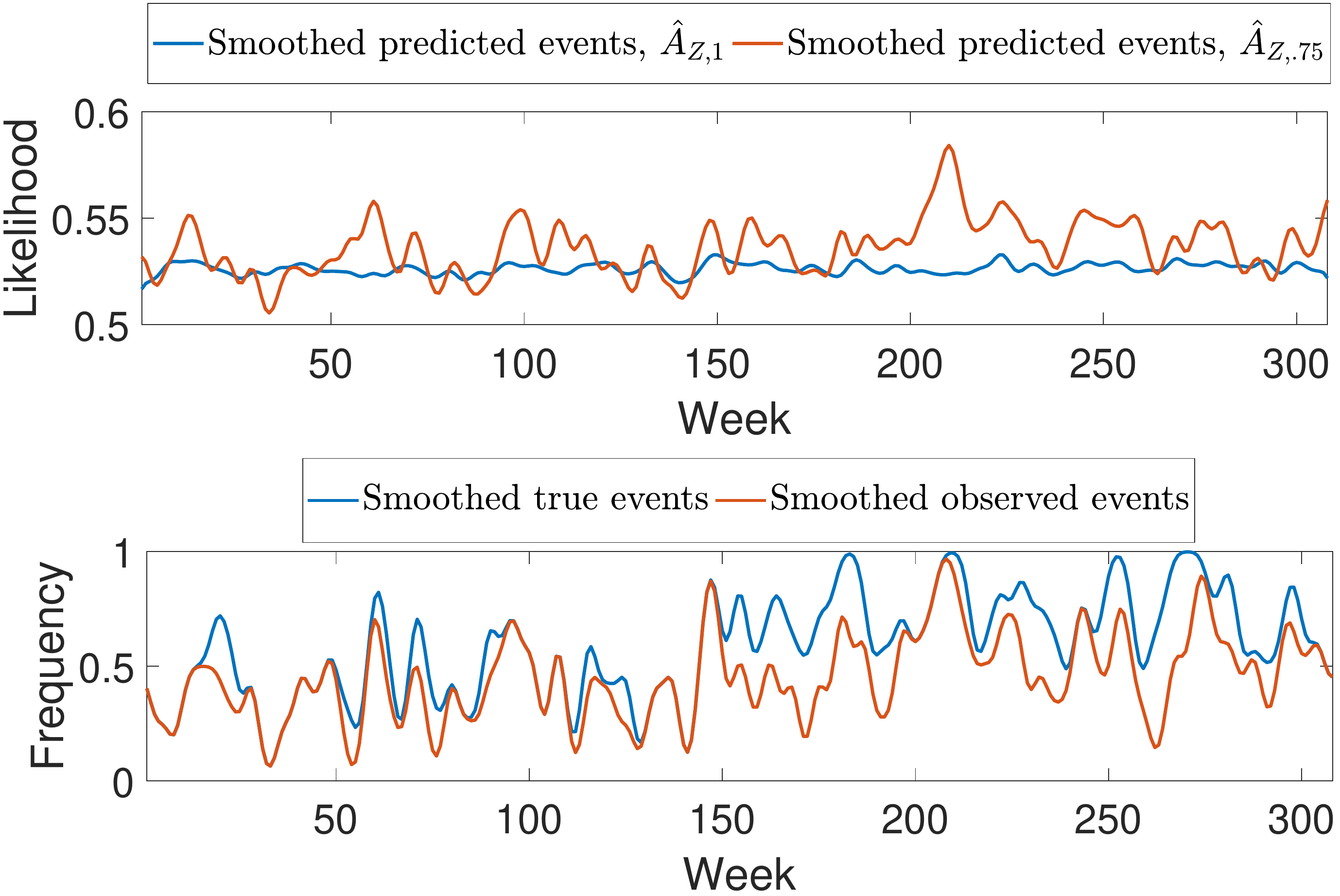}
\caption{\small Result of density propagation on Chicago crime data for Community Area 25 (Austin). After a training period used to estimate $\hat A_{Z,.75}$ (proposed estimator) and $\hat A_{Z,1}$ (naive estimator that doesn't account for missing data), density propagation is run in subsequent test weeks to predict the likelihood of each community area having a homicide at time $n$ based on observations up to time $n-1$. The top panel shows the predicted likelihood of a homicide occurring in the Austin community area of Chicago.  The network $\hat A_{Z,.75}$ predicts $960$ total homicides in the nine community areas during the test period, compared to $748$ for $\hat A_{Z,1}$.  The actual number of homicides was $1035$. The bottom panel shows the true events as well as the partially observed events (after Gaussian smoothing used for visualization).  
}
\label{figure:stochastic_filter}
\end{figure}

Figure \ref{figure:stochastic_filter} displays these likelihoods for Community Area 25 (Austin) which has the largest number of homicides recorded during the test period.  We use Gaussian smoothing to help visualize the trends.  The top panel shows the predicted probability of events using $\hat A_{Z,.75}$ (in red) and the scaled predicted probability of events using $\hat A_{Z,1}$ (in blue).  The bottom panel shows the actual and incompletely observed events during the test period.  The true events generally peak at times in which the predicted events for $\hat A_{Z,.75}$ peak.  For example, both the predicted event and true event charts have peaks around weeks $60$ and $210$.  In contrast, the predicted events for $\hat A_{Z,1}$ are nearly constant over time.  Since it does not account for the missing data (except via a uniform scaling factor), the network $\hat A_{Z,1}$ is not able to capture the dynamics of the process and so it cannot predict events with as much precision as $\hat A_{Z,.75}$. 

\section{Conclusion}
We propose a novel estimator for Bernoulli autoregressive models which accounts for  partially-observed event data. This model can be used in a variety of contexts in which observed discrete event data exhibits autoregressive structure. We provide mean squared error bounds which suggest that our method can accurately capture a network's structure in the presence of missing events.  Simulations and a real data experiment show that our method yields significant improvement compared with ignoring the missed events and that our method is robust to misspecification of the proportion of missed events.  The framework described in this paper suggests a strategy for addressing regression problems with corrupted data in other GLMs, although further work is needed to extend our theoretical analysis beyond binary observations.

\newpage

\appendix
\onecolumn
\section*{Supplementary Material}
The supplementary material is organized as follows.  In Section \ref{sec:supplement_notation} we recall all the notation needed for the proofs.  In Section \ref{sec:statistical_results} we restate the main statistical results
from the paper
and in Section \ref{sec:statistical_proofs} we provide proofs of these results.  In Section \ref{sec:optimization_results} we state the optimization results and give proofs in Section \ref{sec:optimization_proofs}.

\section{Notation and Preliminary Remarks}
\label{sec:supplement_notation}
\begin{itemize}
\item $L_X (a_m):$  negative log likelihood of complete data $X$ given $m$th row $a_m$.
\item $L_X^{(q)}(a_m)$: degree-$q$ Taylor series approximation of $L_X (a_m)$.  
\item $L_Z (a_m):$ negative log likelihood of missing data $Z$ given $m$th row $a_m$.  Loss function is unbiased in the sense that $\mathbb{E}[L_Z (a_m)|X]=L_X (a_m)$. 
\item $L_Z^{(q)}(a_m)$ degree-$q$ Taylor series approximation of $L_Z (a_m)$
\item $R^{(q)}(a_m)=L_X(a_m)-L_Z^{(q)}(a_m)$
\item $\mathbb{B}_{1,\infty}(1)=\{A \in \mathbb{R}^{M \times M} : \|a_m\|_1 \leq 1 \text{ for all } m\}$
\item $p$: fraction of data which is observed
\item $\rho$: $\max_m \|a_m^\ast\|_0$
\end{itemize}

Finally, we introduce additional notation which will be helpful in the proofs of Lemmas \ref{lemma:truncation_z} and \ref{lemma:error}.  First let $\mathcal{U}_d$ denote the set of all monomials of degree $d$.  We represent an element $U \in {\cal U}_d$ as a list containing $d$ elements.  An element in the list corresponds to the index of a term in the monomial (the list can potentially have repeated elements).  For an example, the monomial $x_1^2x_3$ can be represented as the list $(1,1,3)$.  

For a polynomial function $h$ we let $c_{U,h}$ denote the coefficient of the monomial $U$ in $h$.  Finally we define the order of a list to denote the number of unique elements in the list, so $|(1,2)|=2$ whereas $|(1,1)|=1$.    

\paragraph{Example}\textit{Consider the function $h(x_1,x_2)=x_1^2+4x_1x_2$.  Using all the notation above, we can decompose $h$ as 
$$h(x_1,x_2)=\sum_{U \in \mathcal{U}_2} c_{U,h} \prod_{u \in U} x_u$$ where $\mathcal{U}_2=\{(1,1), (1,2),(2,2)\}$ with corresponding coefficients $c_{(1,1),h}=1$, $c_{(1,2),h}=4$ and $c_{(2,2),h}=0$.}

\begin{remark} We next make several observations by applying the notation above to functions which appear in the likelihoods $L_X$ and $L_Z$.  First, for a fixed $t,m$ we decompose the following function as a sum of monomials:
\begin{align*}h(a_{m,1}X_{t,1},\ldots, a_{m,M}X_{t,M}) \vcentcolon =& (a_{m,1}X_{t,1}+\ldots+a_{t,M}X_{t,M})^d \\=& \sum_{U \in \mathcal{U}_{d}} c_{U,h} \left(\prod_{u \in U}a_{m,u}\right) \left(\prod_{u \in U}X_{t,u}\right)\end{align*}
and note that
\begin{equation} \sum_{U \in U_d} c_{U,h} \prod_{u \in U} a_{m,u}=(a_{m,1}+\ldots +a_{m,M})^d \label{eq:helpful1} \end{equation}
We also have 
\begin{align*}\nabla_{a_{m,j}} h=& d \cdot X_{t,j}(a_{m,1}X_{t,1}+\ldots +a_{m,M}X_{t,M})^{d-1} \\=&  \sum_{U \in \mathcal{U}_{d-1}} c_{U,\nabla_{a_{m,j}} h}X_{t,j} \left(\prod_{u \in U}a_{m,u}\right) \left(\prod_{u \in U}X_{t,u}\right)\end{align*}
and we similarly note that 
\begin{equation}\sum_{U \in \mathcal{U}_{d-1}} c_{U,\nabla_{a_{m,j}} h} \prod_{u \in U}a_{m,u}=d \cdot (a_{m,1}+\ldots +a_{m,M})^{d-1} \label{eq:helpful2} \end{equation}
Next consider the function $g$ which appears in the missing data likelihood $L_Z$.
$$g(a_{m,1}Z_{t,1},\ldots, a_{m,M}Z_{t,M}) \vcentcolon = \sum_{U \in \mathcal{U}_{d}} c_{U,g} \left(\prod_{u \in U}a_{m,u}\right) \left(\prod_{u \in U}Z_{t,u}\right)$$ where $c_{U,g}=\frac{c_{U,h}}{p^{|U|}}$.  This observation will be important for our analysis because it allows us to leverage Equation \ref{eq:helpful1}.  Similarly we have 
$$\nabla_{a_{m,j}} g =  \sum_{U \in \mathcal{U}_{d-1}} c_{U,\nabla g_{a_{m,j}}}Z_{t,j} \left(\prod_{u \in U}a_{m,u}\right) \left(\prod_{u \in U}Z_{t,u}\right)$$
where $c_{U,\nabla_{a_{m,j}} g}=\frac{c_{U,\nabla_{a_{m,j}} h}}{p^{|U|}}$ allowing us to use Equation \ref{eq:helpful2}.  

\end{remark}

\begin{remark}
\label{remark:technical}
Using an identical argument to Lemma \ref{lemma:truncation_z} one can show that for $a_m \in \mathbb{B}_1(1)$, $|L_{Z,p}(a_m)-L_{Z,p}^{(q)}(a_m)| \leq (p \pi)^{-q}$.  This implies that $\lim_{q \to \infty} L_{Z,p}^{(q)}$ converges uniformly on $\mathbb{B}_1(1)$ so that $L_{Z,p}(a_m)$ is well defined on this ball.  Moreover, it implies that $\lim_{q \to \infty} \mathbb{E}\left[|L_{Z,p}^{(q)}(a_m)| \Big|X\right]$ converges and so $$\mathbb{E}[L_{Z,p}(a_m)|X]=\lim_{q \to \infty} \mathbb{E}[L_{Z,p}^{(q)}(a_m)|X]=\lim_{q \to \infty} L_X^{(q)}(a_m)=L_X(a_m)$$
and $L_{Z,p}(a_m)$ satisfies \eqref{eq:expectation_property}

\end{remark}

\section{Statistical Results} 
\label{sec:statistical_results}
We assume $A^\ast \in \mathbb{B}_{1,\infty}(1)$ and $p \geq \frac{1}{\pi}$.  We take $q \in \mathbb{N} \cup \{\infty\}$.

\begin{theorem}[Accuracy of $L_Z^{(q)}$]

Suppose $\widehat{A} \in \argmin_{A \in \mathbb{B}_{1,\infty}(1)} L_Z^{(q)}(A)+\lambda \|A\|_1$ where $\lambda \asymp \frac{\log(MT)}{\sqrt{T} (p\pi-1)}+\frac{1}{(p\pi)^{q}}$.  Then $$\|\widehat{A}-A^\ast\|_F^2 \lesssim \frac{s\log^2(MT)}{T(\pi p-1)^2}+\frac{s}{(p\pi)^{2n}}$$ for $T \gtrsim \rho^2 \log(MT)$ with probability at least $1-\frac{1}{T}$.  
\label{thm:main}
\end{theorem}

The proof of Theorem \ref{thm:main} relies on the following supplementary lemmas.

\begin{lemma}
Let $f(x)=\log(1+\exp(x))$.  Then $|\frac{f^{(q)}(0)}{q!}| \lesssim \frac{1}{q\pi^{q}}$. 
\label{lemma:prelim}
\end{lemma}

\begin{lemma}[Truncation error of $\nabla L_X$]
Suppose $\|a_m\|_1 \leq 1$.  Then $$\|\nabla L_X(a_m)-\nabla L_X^{(q)}(a_m)\|_\infty \lesssim \pi^{-q}.$$
\label{lemma:truncation_x}
\end{lemma}

\begin{lemma}[Truncation Error of $\nabla L_Z$] Suppose $\|a_m\|_1 \leq 1$.  Then $$\|\nabla L_Z(a_m)- \nabla L_Z^{(q)}(a_m)\|_\infty \lesssim (p\pi)^{-q}.$$
\label{lemma:truncation_z}
\end{lemma}

\begin{lemma}
$$\sup_{\|a_m\|_1 \leq 1} \left\| \nabla L_Z^{\log(T)}(a_m)- \nabla L_X^{\log(T)}(a_m)\right\|_\infty \lesssim \frac{\log(MT)}{\sqrt{T} (p\pi-1)}$$ with probability at least $1-\frac{\log(T)^2}{MT^2}$.
\label{lemma:error}
\end{lemma}

\begin{lemma}
$$\sup_{\|a_m\|_1 \leq 1} \left\| \nabla L_Z^{(q)}(a_m)- \nabla L_X(a_m)\right\|_\infty \lesssim \frac{\log(MT)}{\sqrt{T} (p\pi-1)}+\frac{1}{(p\pi)^{q}}$$ with probability at least $1-\frac{\log(T)^2}{MT^2}$.
\label{lemma:error2}
\end{lemma}

\begin{lemma}
Let $\|v\|_T^2=\frac{1}{T}\sum_t (v^\top X_t)^2$ and $\tilde{R}=\min(R_{\min},1-R_{\max})$.  For any $v \in \mathbb{R}^M$ we have 
$$\|v\|_T^2 \geq \frac{\tilde{R}}{2}\|v\|_2^2-\sqrt{\frac{3\log(M)}{T}}\|v\|_1^2$$ and 
$$\|v\|_T^2 \leq \frac{1}{4}\|v\|_2^2+\sqrt{\frac{3\log(M)}{T}}\|v\|_1^2$$ with probability at least $1-\frac{1}{M}$.
\label{lemma:rsc}
\end{lemma}

\section{Proofs of Statistical Results}
\label{sec:statistical_proofs}

\subsection{Proof of Theorem \ref{thm:main}}
\paragraph{Part 1: Controlling the Remainder}
We set $\lambda \asymp \frac{\log(MT)}{\sqrt{T} (p\pi-1)}+\frac{1}{(p\pi)^{q}}$ and let $\triangle_m=\widehat{a}_m-a_m^\ast$.  Note that the loss functions are decomposable, i.e., $L_Z(A)=\sum_m L_Z (a_m)$.  Since $\widehat{A} \in \argmin_{A \in \mathbb{B}_{1,\infty}(1)} L_Z^{(q)}(A)$ we have 
$$L_Z^{(q)}(\widehat{a}_m) \leq L_Z^{(q)}(a_m^\ast)$$ and so $$L_X(\widehat{a}_m) \leq L_X(a_m^\ast)+\left(R^{(q)}(\widehat{a}_m)-R^{(q)}(a_m^\ast)\right).$$ 
Define $\triangle_m:=\widehat{a}_m-a_m^\ast$.  By the mean value theorem, there exists some $v \in \mathbb{B}_1(1)$ such that $$R^{(q)}(\widehat{a}_m)-R^{(q)}(a_m^\ast)=\langle \nabla R^{(q)}(v),\triangle_m \rangle$$ so by Lemma \ref{lemma:error2}
$$|R^{(q)}(\widehat{a}_m)-R^{(q)}(a_m^\ast)| \leq \|\triangle_m\|_1 \|\nabla R^{(q)}(v)\|_\infty \leq  \frac{\lambda}{4}\|\triangle_m\|_1$$ with probability $1-\frac{\log(T)^2}{MT^2}$. We condition on this event for the remainder of the proof. 

\paragraph{Part 2: Setting Up the Standard Equations}
The next several steps follow standard techniques for $\ell_1$ regularization in GLMs.  Expanding $L_X$ and using the substitution  $$X_{t+1,m}=\mathbb{E}[X_{t+1,m}|X_t]+\epsilon_{t,m}=f'(a_m^{\ast\top} X_t)+\epsilon_{t,m}$$
gives 
\begin{align*}\frac{1}{T} \sum_t f(\widehat{a}_m^\top X_t&)-(\widehat{a}_m^\top X_t)(f'(a_m^{\ast\top}X_t) +\epsilon_{t,m})+\lambda \|\widehat{a}_m\|_1\\
&\leq \frac{1}{T} \sum_t f(a_m^{\ast\top} X_t)-(a_m^{\ast\top} X_t)(f'(a_m^{\ast\top}X_t) +\epsilon_{t,m})  + \lambda \|a_m^\ast\|_1+\frac{\lambda}{4}\|\triangle_m\|_1
\end{align*}
Rearranging terms yields
\begin{align} \label{eq:top}\frac{1}{T}\sum_t f(\widehat{a}_m^\top X_t)&-f(a_m^{\ast\top}X_t)-f'(a_m^{\ast\top}X_t)\triangle_m^\top X_t \\ &\leq |\frac{1}{T}\sum_t \epsilon_{t,m}\triangle_m^\top X_t| + \lambda\left(\|a_m^\ast\|_1-\|\widehat{a}_m\|_1\right)+\frac{\lambda}{4}\|\triangle_m\|_1 \notag \end{align}
Since $f(x)=\log(1+\exp(x))$ is $\sigma$-strongly convex on $[R_{\min},R_{\max}]$ with $\sigma=\frac{R_{min}}{(1+R_{\min})^2}$ we can lower bound Equation \eqref{eq:top} by $\frac{\sigma}{T}\sum_t (\triangle_m^\top X_t)^2$.  
Also note 
$$\left|\frac{1}{T}\sum_t \epsilon_{t,m}\triangle_m^\top X_t\right| \leq \left \|\triangle_m\right \|_1 \left \|\frac{1}{T} \sum_t \epsilon_{t,m}X_t\right \|_\infty.$$  Using Theorem 2.2 in \cite{hall2016inference}, $\left \|\frac{1}{T} \sum_t \epsilon_{t,m}X_t\right \|_\infty \leq \frac{\lambda}{4}$ with probability at least $1-\frac{1}{MT}$.  Applying these observations to Equation \eqref{eq:top} gives
\begin{equation*}\frac{1}{T}\sum_{t=1}^T (\triangle_m^\top X_t)^2 \leq \frac{\lambda}{2}\|\triangle_m\|_1+\lambda \|a_m^\ast\|_1-\lambda \|\widehat{a}_m\|_1. \end{equation*} Define $S:=\{i :a_{m,i}^\ast \not =0\}$ and $\rho_m:= \|a_m^\ast\|_0$.  Then \begin{equation}\frac{1}{T}\sum_{t=1}^T (\triangle_m^\top X_t)^2 \leq \frac{3\lambda}{2} \|\triangle_{m,S}\|_1 - \frac{\lambda}{2} \|\triangle_{m,S^c}\|_1\label{eq:initial_eq}\end{equation}
and so
\begin{align}\frac{1}{T}\sum_{t=1}^T (\triangle_m^\top X_t)^2  \leq & \frac{3\lambda}{2} \|\triangle_{m,S}\|_1 
\leq  \frac{3\lambda \sqrt{\rho_m}}{2} \|\triangle_{m,S}\|_2 \leq \frac{3\lambda \sqrt{\rho_m}}{2} \|\triangle_{m}\|_2.\label{eq:second}\end{align}
\paragraph{Restriction of $\triangle_m$ to Cone}It remains to lower bound $\frac{1}{T} \sum_t (\triangle_m^\top X_t)^2$ in terms of $\|\triangle_m\|_2^2$.  In order to do this we will rely heavily on the fact that $\triangle_m$ is not an arbitrary vector.  Instead we show that $\triangle_m$ must lie in a cone with important properties.  In particular, returning to Equation \eqref{eq:initial_eq} and observing that $0 \leq \frac{1}{T} \sum_t (\triangle_m^\top X_t)^2$ it follows that $\|\triangle_{m,S^c}\|_1 \leq 3 \|\triangle_{m,S}\|_1$.  Thus \begin{equation}\|\triangle_m\|_1=\|\triangle_{m,S}\|_1+\|\triangle_{m,S^c}\|_1 \leq 4\|\triangle_{m,S}\|_1. \label{eq:cone}\end{equation}
\paragraph{Restricted Eigenvalue Condition}
As mentioned in the previous section, our goal is to lower bound $\frac{1}{T}\sum_{t=1}^T (\triangle_m^\top X_t)^2$.  This is commonly referred to as a restricted eigenvalue condition in the literature, and it is closely related to the restricted strong convexity condition proved in Lemma \ref{lemma:rsc/rsm}.  In particular, Lemma \ref{lemma:rsc} guarantees that there exist universal constants $c_1$ and $c_2$ such that 
\begin{align*}\frac{1}{T} \sum_t (\triangle_m^\top X_t)^2 \geq \frac{c_1}{2}\|\triangle_m\|_2^2-c_2 \sqrt{\frac{\log(MT)}{T}} \|\triangle_m\|_1^2.\end{align*}
For arbitrary vectors this lower bound can be negative; however, by Equation \eqref{eq:cone} 
$$\|\triangle_m\|_1^2 \leq 16\|\triangle_{m,S}\|_1^2 \leq 16\rho_m \|\triangle_{m,S}\|_2^2 \leq 16\rho_m \|\triangle_m\|_2^2$$ and thus 
$$\frac{1}{T} \sum_t (\triangle_m^\top X_t)^2 \geq \frac{c_1}{2} \|\triangle_m\|_2^2-16c_2\rho_m \sqrt{\frac{\log(MT)}{T}} \|\triangle_m\|_2^2.$$
Hence if $T \gtrsim \rho_m^2 \log(MT)$ it follows that 
$$\frac{1}{T} \sum_t (\triangle_m^\top X_t)^2 \geq c\|\triangle_m\|_2^2$$ for a universal constant $c$.  Plugging this in to Equation \eqref{eq:second} gives that 
$$\|\triangle_m\|_2 \lesssim \sqrt{\rho_m}\lambda$$ with probability $1-\frac{1}{MT}$.  Taking a union bound we conclude that 
$$\|A^\ast-\widehat{A}\|_F^2 = \sum_{m=1}^M \|\triangle_m\|_2^2 \lesssim \sum_{m=1}^M \rho_m \lambda^2 = s\lambda^2$$
with probability $1-\frac{1}{T}$.  \qed

\subsection{Proof of Lemma \ref{lemma:prelim}}
A computation shows that 
$$f^{(q)}(0)=\frac{1}{2^{q}} \sum_{m=0}^{q-2} (-1)^{m} A(q-1,m)$$
 where the $A(q,m)$ are the Eulerian numbers.  The alternating sum of the Eulerian numbers for fixed $q$ can be given as 
$$\sum_{m=0}^{q-2} (-1)^{q} A(q-1,m)=\frac{2^{q}(2^{q}-1)B_q}{q}$$
 where $B_q$ is the $q$th Bernoulli number (see the derivation of Equation 4.8 in \cite{euclerian_numbers}).  Thus $$\left|\frac{f^{(q)}(0)}{q!}\right|\leq \frac{2^{q}|B_q|}{q!q}.$$  Using \cite{alzer} we have the bound $|B_q| \lesssim \frac{q!}{(2\pi)^{q}}$.  Thus $$\left|\frac{f^{(q)}(0)}{q!}\right| \lesssim \frac{2^{q}q!}{q!q2^{q}\pi^{q}} = \frac{1}{q\pi^{q}}.$$ \qed

\subsection{Proof of Lemma \ref{lemma:truncation_x}}
For any $j$ we have 
\begin{align*} \left|(\nabla L_X(a_m)-\nabla L_X^{(q)}(a_m))_j\right| = &\left|\frac{1}{T} \sum_t \sum_{i=q}^\infty \frac{f^{(i)}(0)}{(i-1)!} (a_m^T X_t)^{i-1}X_{t,j}\right|\\
\leq & \frac{1}{T} \sum_t \sum_{i=q}^\infty \pi^{-i} \\
\lesssim & \frac{1}{T} \sum_t \pi^{-q}=\pi^{-q},
\end{align*}
where the last two lines use Lemma \ref{lemma:prelim} along with the fact that $\|a_m\|_1 \leq 1$. \qed

\subsection{Proof of Lemma \ref{lemma:truncation_z}}
Differentiating $L_Z(a_m)$ with respect to $a_{m,j}$ gives 
$$\nabla_{j} L_Z(a_m) = \sum_{d=1}^\infty \left(\frac{f^{(d)}(0)}{d!T} \sum_t \sum_{U \in \mathcal{U}_{d-1}} \frac{c_{U,\nabla_j f}}{p^{|U|}} \left(\prod_{u \in U}a_{m,u}\right) \left(\prod_{u \in U}Z_{t,u}\right)\right)$$
We first bound 
$$g_d(a_m):=\frac{f^{(d)}(0)}{d!T} \sum_t \sum_{U \in \mathcal{U}_{d-1}} \frac{c_{U,\nabla_j f}}{p^{|U|}} \left(\prod_{u \in U}a_{m,u}\right) \left(\prod_{u \in U}Z_{t,u}\right)$$
which is the degree $d-1$ term of $\nabla_j L_Z(a_m)$.  All the terms other than the $a_{m,u}$ in $g_d(a_m)$  are always non-negative and $|U| \leq d$, so 
$$ |g_d(a_m)| \leq \frac{f^{(d)}(0)}{d!p^{d}} \frac{1}{T}\sum_t   \sum_{U \in \mathcal{U}_{d-1}} c_{U,\nabla_j f} \prod_{u \in U}|a_{m,u}|.$$
By Equation \ref{eq:helpful2}
$$\sum_{U \in \mathcal{U}_{d-1}} c_{U,\nabla_j f} \prod_{u \in U}|a_{m,u}|=d(|a_{m,1}|+\ldots +|a_{m,M}|)^{d-1} \leq d.$$
We conclude that $$|g_d(a_m)|  \leq \frac{1}{T}\sum_t \frac{f^{(d)}(0)}{(d-1)!p^{d}}= \frac{f^{(d)}(0)}{(d-1)!p^{d}}.$$  Using Lemma \ref{lemma:prelim} $$\frac{f^{(d)}(0)}{(d-1)!p^{d}} \lesssim \frac{1}{(p\pi)^{d}}.$$ 

Overall, we have concluded that $|g_d(a_m)| \leq \frac{1}{(p\pi)^{d}}$.  We are ultimately interested in $|\nabla_j L_Z(a_m)-\nabla_j L_Z^{(q)}(a_m)|$.  This is the sum of the degree $d$ terms of $\nabla_j L_Z(a_m)$ for all $d \geq q$.  In other words,
$$|\nabla_j L_Z(a_m)-\nabla_j L_Z^{(q)}(a_m)| \leq \sum_{d=q}^\infty |g_d(a_m)| \leq \sum_{d=q}^\infty \frac{1}{(p\pi)^d} \lesssim \frac{1}{(p\pi)^{q}}$$ as claimed. \qed

\subsection{Proof of Lemma \ref{lemma:error}} 
We begin by bounding individual monomials of $\left\| \nabla L_Z^{\log(T)}(a_m)- \nabla L_X^{\log(T)}(a_m)\right\|_\infty$.  We then extend these individual bounds to bounds on the entire expression.
\paragraph{Bounding Individual Monomials}
 Following the notation introduced in Section \ref{sec:supplement_notation} a degree $d-1$ monomial of $ \nabla_j L_X^{\log(T)}(a_m)$ with index $U$ is of the form 
 $$m_{U,X}:=\frac{f^d(0)}{d!T} \sum_t c_{U,\nabla_j f} \left(\prod_{u \in U} a_{m,u}\right) \left(\prod_{u \in U}X_{t,u}\right).$$Meanwhile, the degree $d-1$ monomial of $ \nabla_j L_Z^{\log(T)}(a_m)$ with index $U$ is of the form
$$m_{U,Z}:=\frac{f^d(0)}{d!Tp^{|U|}} \sum_t c_{U,\nabla_j f}\left(\prod_{u \in U} a_{m,u}\right) \left(\prod_{u \in U}Z_{t,u}\right).$$
The difference of these monomials is given by 
\begin{align*}&|m_{U,X}-m_{U,Z}| \\= & \Bigg|\frac{f^d(0)}{Td!} \Big(\sum_t c_{U,\nabla_j f} \prod_{u \in U} a_{m,u}\prod_{u \in U}X_{t,u}- \frac{1}{p^{|U|}} \sum_t c_{U,\nabla_j f} \prod_{u \in U} a_{m,u}\prod_{u \in U}Z_{t,u}\Big)\Bigg|\\
=& \frac{f^d(0) c_{U,\nabla_j h}\prod_{u \in U}|a_{m,u}|}{Td!} \left|\left(\sum_t \prod_{u \in U} X_{t,u}-\frac{1}{p^{|U|}} \sum_t \prod_{u \in U} Z_{t,u}\right)\right|.
\end{align*}
Observe that $$ \frac{1}{p^{|U|}}\prod_{u \in U} Z_{t,u} \in \{0,\frac{1}{p^{|U|}}\}$$ and $$\mathbb{E}\left[\frac{1}{p^{|U|}}\prod_{u \in U} Z_{t,u}\right]=\prod_{u \in U} X_{t,u}.$$  We apply Hoeffding's inequality to conclude that 
\begin{align} \mathbb{P}&\left( \left|\sum_t \frac{1}{p^{|U|}}\prod_{u \in U} Z_{t,u}-\sum_t\prod_{u \in U} X_{t,u}\right| \geq \frac{\log(MT) \sqrt{T}}{p^{|U|}} \right) \leq  2\exp(-2\log^2(MT)).\label{eq:concentration} \end{align} 
Thus
\begin{align}
    |m_{U,X}-m_{U,Z}| \leq \frac{f^d(0) \log(MT) c_{U,\nabla_j h} \prod_{u \in U}|a_{m,u}|}{d!\sqrt{T}p^{|U|}} \label{eq:single_monomial}
\end{align}
with probability at least $1-2\exp(-2\log^2(MT))$.

\paragraph{Extension to Entire Expression}
We need to take a union bound so that this holds for all monomials of degree at most $\log(T)$.  However, since $Z_{t,u}$ and $X_{t,u}$ are binary random variables
$$\sum_t \prod_{u \in U} Z_{t,u} - \sum_t \prod_{u \in U} X_{t,u}= \sum_t Z_{t,v}\prod_{u \in U} Z_{t,u} - \sum_t X_{t,v}\prod_{u \in U} X_{t,u}$$ whenever $v \in U$.  

Suppose we have shown that Equation \eqref{eq:concentration} holds for all monomials of degree $<d$.
Now to show it holds for all monomials of degree $d$ we only need to show that  \eqref{eq:concentration} holds for all monomials of degree $d$ that have $d$ distinct terms.  The remaining concentrations are already covered by the degree $d-1$ monomials.  There are $\binom{M}{d} \leq M^d$ monomials of degree $d$ with $d$ distinct terms. Hence we need to take a union bound over at most $$\sum_{d=1}^{\log(T)} M^d \leq \log(T)M^{\log(T)}$$ monomials of degree $\leq \log(T)$ and so  
\begin{align*} \mathbb{P}\Bigg( \Big| \sum_t \frac{1}{p^{|U|}}\prod_{u \in U} Z_{t,u}-\sum_t&\prod_{u \in U} X_{t,u}\Big| \geq \frac{\log(MT) \sqrt{T}}{p^{|U|}}  \text{ for } \geq 1 \text{ monomial of degree} \leq \log(T)\Bigg)\\& \leq  2\log(T)\exp\left(\log(M)\log(T)-2\log^2(MT)\right)\leq \frac{1}{M^2T^2}. \end{align*}  We condition on this event and recall that $\mathcal{U}_{d-1}$ denotes the set of all monomials of degree $d-1$. Using Equation \eqref{eq:single_monomial}, the difference between the degree $d-1$ terms of $\nabla_j L_X^{\log(T)}(a_m)$ and $\nabla_j L_Z^{\log(T)}(a_m)$ can be bounded by 
$$\sum_{U \in \mathcal{U}_{d-1}} |m_{U,X}-m_{U,Z}| \leq \frac{f^d(0)\log(MT)}{d! \sqrt{T}} \sum_{U \in \mathcal{U}_{d-1}} \frac{c_{U,\nabla_j f }}{p^{|U|}} \prod_{u \in U} |a_{m,u}|.$$ 
Using Lemma \ref{lemma:prelim} and Equation \ref{eq:helpful2} along with the fact that $|U| \leq d$, 
\begin{align*}\frac{f^d(0)\log(MT)}{d! \sqrt{T}} \sum_{U \in \mathcal{U}_{d-1}} \frac{c_{U,\nabla_j f }}{p^{|U|}} \prod_{u \in U} |a_{m,u}| \leq & \frac{f^d(0)\log(MT)}{(d-1)!p^d \sqrt{T}} \\ \leq & \frac{\log(MT)}{\sqrt{T} (p\pi)^d}\end{align*} where the final inequality uses Lemma \ref{lemma:prelim}.
Thus we have the bound
\begin{align*}\left|\nabla_j L_X^{\log(T)}(a_m)- \nabla_j L_Z^{\log(T)}(a_m)\right| \leq & \sum_{d=1}^{\log(T)}  \frac{\log(MT)}{\sqrt{T} (p\pi)^d} \\ \lesssim & \frac{\log(MT)}{\sqrt{T} (p\pi-1)}\end{align*} with probability at least $1-\frac{\log(T)^2}{MT^2}$. \qed

\subsection{Proof of Lemma \ref{lemma:error2}}
By the triangle inequality we have 
\begin{align*}\left|\left(\nabla L_Z^{(q)}(a_m)- \nabla L_X(a_m)\right)_j\right|& \leq \\  &\left|\left(\nabla L_Z^{(q)}(a_m)- \nabla L_Z^{\log(T)}(a_m)\right)_j\right|\\ +&\left|\left(\nabla L_X^{\log(T)}(a_m)- \nabla L_X(a_m)\right)_j\right| \\ +&  \left|\left(\nabla L_Z^{\log(T)}(a_m)- \nabla L_X^{\log(T)}(a_m)\right)_j\right|.\end{align*}
By Lemmas \ref{lemma:truncation_x} and \ref{lemma:truncation_z} the first two terms can be bounded by $\max\left((p\pi)^{-\log(T)},(p\pi)^{-q}\right)$  while by Lemma \ref{lemma:error} the final term can be bounded is $$\lesssim \frac{\log(MT)}{\sqrt{T} (p\pi-1)}$$ with probability at least $1-\frac{\log(T)^2}{MT^2}$.  We conclude that $$\|\nabla R^{(q)}(a_m)\|_\infty \lesssim \frac{\log(MT)}{\sqrt{T} (p\pi-1)}+\frac{1}{(p\pi)^{-q}}$$ with probability at least $1-\frac{\log(T)^2}{MT^2}$ as claimed. \qed

\subsection{Proof of Lemma \ref{lemma:rsc}}
We have \begin{align}\|v\|_T^2= \frac{1}{T} \sum_t v^\top \left(\mathbb{E}[(X_tX_t^\top)|X_{t-1}]\right) v - \frac{1}{T} \sum_t v^\top \left(X_tX_t^\top -\mathbb{E}[X_tX_t^\top|X_{t-1}]\right)v\label{eq:rsc1}\end{align}
By Theorem 2.1 in \cite{hall2016inference}, 
\begin{equation}\frac{1}{T} \sum_t v^\top \left(\mathbb{E}[(X_tX_t^\top)|X_{t-1}]\right) v \geq \frac{\tilde{R}}{2}\|v\|_2^2\label{eq:rsc2}\end{equation}
Now we define the matrix $G \in \mathbb{R}^{M \times M}$ as follows:
$$
G := \frac{1}{T}\sum_{t \in \mathcal{T}}(X_tX_t^\top - \mathbb{E}[X_tX_t^\top|X_{t-1}]).
$$
Note that each entry of $G$ is a martingale and $$v^\top \left(\sum_t X_tX_t^\top -\mathbb{E}[X_tX_t^\top|X_{t-1}]\right)v \leq \|v\|_1^2 \max_{m,m'}|G_{m,m'}|.$$
Applying the Azuma-Hoeffding inequality we conclude that for any $m,m'$
$$\mathbb{P}\left(|G_{m,m'}| \geq \sqrt{3\log(M)T}\right) \leq \frac{1}{M^3}$$ and therefore 
$$\mathbb{P}\left(\max_{m,m'}|G_{m,m'}| \geq \sqrt{3\log(M)T}\right) \leq \frac{1}{M}.$$ 
Overall we have concluded 
\begin{align} \frac{1}{T}  \sum_t v^\top \left(X_tX_t^\top -\mathbb{E}[X_tX_t^\top|X_{t-1}]\right)v \leq \frac{\sqrt{3\log(M)}}{\sqrt{T}}\|v\|_1^2 \label{eq:rsc3} \end{align} with probability at least $1-\frac{1}{M}$.  Combining Equations \eqref{eq:rsc1},\eqref{eq:rsc2} and \eqref{eq:rsc3} give the desired result. \qed

\newpage

\section{Optimization Results}
\label{sec:optimization_results}

The main result of this section is the following Theorem.

\begin{theorem} 
Suppose $A^\ast \in \mathbb{B}_{1,\infty}(1)$ and $\|a_m^\ast\|_0>0$ for at least $\frac{M}{C}$ rows of $A^\ast$ where $C$ is a universal constant.  Let $\tilde{A} \in \mathbb{B}_{1,\infty}(1)$ be any stationary point of $L_Z^{(q)}(A)+\lambda \|A\|_1$ where $\lambda \asymp \frac{\log(MT)}{\sqrt{T} (p\pi-1)}+\frac{1}{(p\pi)^{q}}$.  Then 
$$\|\tilde{A}-A^\ast\|_F^2 \lesssim s \left(\sqrt{\frac{\log(MT)}{T}}+\frac{1}{(p\pi)^{q}}\right) $$
with probability at least $1-\frac{\log(T)}{T^2}$.
\label{thm:main_opt}
\end{theorem}

In order to prove Theorem \ref{thm:main_opt} we need to introduce notions of Restricted Strong Convexity (RSC) and Restricted Smoothness (RSM) from \cite{fast_global}.  To do this we first define the first order Taylor expansion to a function $L$:
$$T_L(v,w)=L(v)-L(w)-\langle \nabla L(w), v-w\rangle.$$
\begin{definition}[Restricted Strong Convexity]A loss function $L$ satisfies the RSC condition with parameters $\alpha,\tau$ if 
$$T_L(v,w) \geq \frac{\alpha}{2}\|v-w\|_2^2-\tau \|v-w\|_1^2$$ for all $v,w \in \mathbb{B}_1(1)$. 
\end{definition}
\begin{definition}[Restricted Smoothness]
A loss function $L$ satisfies the RSM condition with parameters $\alpha,\tau$ if 
$$T_L(v,w) \geq \frac{\alpha}{2}\|v-w\|_2^2+\tau \|v-w\|_1^2$$ for all $v,w \in \mathbb{B}_1(1)$. 
\end{definition}

\begin{lemma}

The RSC and RSM conditions are satisfied for $L_Z^{(q)}$ with constants 
$\alpha=c_1$ and $\tau=c_2\left(\sqrt{\frac{\log(MT)}{T}}+\frac{1}{(p\pi)^{q}}\right)$ with probability at least $1-\frac{\log(T)}{T^2}$ where $c_1$ and $c_2$ are universal constants.

\label{lemma:rsc/rsm}
\end{lemma}

Combining Lemma \ref{lemma:rsc/rsm} with Theorem 2 in \cite{fast_global} gives the following corollary.

\begin{cor}
Under the conditions of Theorem \ref{thm:main}, let $a_m^s$ denote the $s$'th iteration of projected gradient descent using the loss function $\phi(a_m)=L_Z^{(q)}(a_m)-\lambda \|a_m\|_1$.  There exists some $S$ such that for all $s>S$ we have 
$$\phi(a_m^s)-\phi(\widehat{a}_m) \lesssim \tau \rho_m.$$

\label{cor:loss_bound}
\end{cor}

\newpage

\section{Proofs of Optimization Results}
\label{sec:optimization_proofs}

\subsection{Proof of Lemma \ref{lemma:rsc/rsm}}
We begin by computing the first order Taylor approximation:
\begin{align*}T_{L_Z^{(q)}}&(v,w)\\&=L_Z^{(q)}(v)-L_Z^{(q)}(w)-\langle \nabla L_Z^{(q)}(w),v-w\rangle \\ &=\underbrace{L_X(v)-L_X(w)-\langle \nabla L_X(w),v-w\rangle}_{1} + \underbrace{R^{(q)}(v)-R^{(q)}(w)-\langle \nabla R^{(q)}(w),v-w\rangle}_{2}.\end{align*} We first handle term (1), which is the first order Taylor error for $L_X$.  A computation shows that this is equal to 
$$\frac{1}{T} \sum_T f(v^\top X_t)-f(w^\top X_t)-f'(w^\top X_t) \langle v-w,X_t\rangle$$ where again $f(x)=\log(1+\exp(x))$.  Since $f$ is $\sigma$-strongly convex on $[R_{\min},R_{\max}]$ with $\sigma=\frac{R_{\min}}{(1+R_{\min})^2}$ we can lower bound term (1): 
$$\frac{\sigma}{T} \sum_T \langle v-w,X_t\rangle ^2 \leq L_X(v)-L_X(w)-\langle \nabla L_X(w),v-w\rangle.$$ Using Lemma \ref{lemma:rsc}, term (1) can be bounded below by

and can be bounded above by
\begin{equation} \frac{\sigma}{4}\|v-w\|_2^2+\sigma\sqrt{\frac{3\log(M)}{T}}\|v-w\|_1^2\label{eq:term11}\end{equation}
and below by
\begin{equation} \frac{\sigma\tilde{R}}{2}\|v-w\|_2^2-\sigma\sqrt{\frac{3\log(M)}{T}}\|v-w\|_1^2.\label{eq:term12} \end{equation}
It remains to handle term (2) which is $T_{R^{(q)}}(v,w)$. By the mean value theorem there exists some $u \in \mathbb{B}_1(1)$ such that $R^{(q)}(v)-R^{(q)}(w)=\langle \nabla R^{(q)}(u), v-w \rangle$.  Thus we can bound term (2) by 
$$\left(\|\nabla R^{(q)}(u)\|_\infty+\|\nabla R^{(q)}(w)\|_\infty\right)\|v-w\|_1.$$
By Lemma \ref{lemma:error2}
$$\|\nabla R^{(q)}(u)\|_\infty+\|\nabla R^{(q)}(w)\|_\infty \lesssim \frac{\log(MT)}{\sqrt{T}}+\frac{1}{(p\pi)^{q}}
$$with probability $1-\frac{2\log(T)}{MT^2}$.  Combining this with our bounds on term 1 in Equations \eqref{eq:term11} and \eqref{eq:term12} gives the final result. \qed

\subsection{Proof of Theorem \ref{thm:main_opt}}
By Corollary \ref{cor:loss_bound} we have 
$$\left(L_Z^{(q)}(a_m^s)-L_Z^{(q)}({\widehat{a}_m}\right) +\left (\lambda \|\widehat{a}_m\|_1-\lambda \|a_m^s\|_1\right) \leq \tau \rho_m.$$
Since $\widehat{a}_m$ is a stationary point it satisfies 
$$\langle \nabla L_Z^{(q)} (\widehat{a}_m)-\lambda \widehat{a}_m, a_m^s-\widehat{a}_m \rangle \geq 0.$$
Using this, along with the fact that $a_m^s, \widehat{a}_m \in \mathbb{B}_1(1)$ we get that 
$$T_{L_Z^{(q)}}(a_m^s,\widehat{a}_m) \leq \tau \rho_m + 3\lambda.$$
Using the RSC condition from Lemma \ref{lemma:rsc/rsm} we conclude
$$\|\widehat{a}_m-a_m^s\|_2^2 \lesssim \tau \rho_m + \tau + \lambda \lesssim (\rho_m+1) \left(\sqrt{\frac{\log(MT)}{T}}+\frac{1}{(p\pi)^{q}}\right).$$
Finally we apply the statistical error bound on $\|\widehat{a}_m-a_m^\ast\|_2^2$ from Theorem \ref{thm:main} along with the triangle inequality to conclude that 
$$\|a_m^s-a_m^\ast\|_2^2 \lesssim (\rho_m+1) \left(\sqrt{\frac{\log(MT)}{T}}+\frac{1}{(p\pi)^{q}}\right).$$  Summing over all $m$ and assuming $\|a_m^\ast\|_0>0$ for at least $\frac{M}{C}$ values of $m$ allows us to conclude that 
$$\|A^s-A^\ast\|_F^2 \lesssim s \left(\sqrt{\frac{\log(MT)}{T}}+\frac{1}{(p\pi)^{q}}\right).$$
To get the final form of the result, we recall that $A^s$ is the $s$th iteration of projected gradient descent run with an arbitrary initialization within $\mathbb{B}_{1,\infty}(1)$.  In particular, if we initialize $A^0$ at a stationary point then $A^s=A^0$ which gives the final form of the Theorem.

\end{document}